\pdfoutput=1
\documentclass{article}

    \PassOptionsToPackage{numbers, compress}{natbib}

\usepackage[main, final]{neurips_2026}

\usepackage[utf8]{inputenc} %
\usepackage[T1]{fontenc}    %
\usepackage{hyperref}       %
\usepackage{url}            %
\usepackage{booktabs}       %
\usepackage{amsfonts}       %
\usepackage{nicefrac}       %
\usepackage{microtype}      %
\usepackage{xcolor}         %
\usepackage{caption}
\newcommand{\NA}{\multicolumn{1}{c}{--}}

\usepackage[ruled,vlined,linesnumbered]{algorithm2e}

\usepackage{amsmath,amsfonts,bm}

\def\eqref#1{equation~\ref{#1}}

\def\1{\bm{1}}

\def\rvu{{\mathbf{u}}}

\def\rvx{{\mathbf{x}}}
\def\rvy{{\mathbf{y}}}
\def\rvz{{\mathbf{z}}}

\def\vmu{{\bm{\mu}}}

\def\vu{{\bm{u}}}

\def\mI{{\bm{I}}}

\def\mM{{\bm{M}}}

\def\mU{{\bm{U}}}

\DeclareMathAlphabet{\mathsfit}{\encodingdefault}{\sfdefault}{m}{sl}
\SetMathAlphabet{\mathsfit}{bold}{\encodingdefault}{\sfdefault}{bx}{n}

\def\gC{{\mathcal{C}}}

\def\gN{{\mathcal{N}}}

\def\gP{{\mathcal{P}}}
\def\gQ{{\mathcal{Q}}}

\def\gT{{\mathcal{T}}}

\newcommand{\E}{\mathbb{E}}

\newcommand{\KL}{D_{\mathrm{KL}}}

\title{Learning to Assimilate: Amortized Control for Autoregressive Diffusion}

\author{%
  \textbf{Prakhar Srivastava} \quad
  \textbf{Farrin Marouf Sofian} \quad
  \textbf{Francesco Immorlano} \\
  \textbf{Kushagra Pandey} \quad
  \textbf{Stephan Mandt} \\
  University of California, Irvine \\
  \texttt{\{prakhs2,fmaroufs,fimmorla,pandeyk1,mandt\}@uci.edu}
}

\definecolor{citationmagenta}{RGB}{150,40,105}
\definecolor{referenceblue}{RGB}{30,75,140}
\definecolor{urlteal}{RGB}{0,105,105}
\hypersetup{
    colorlinks=true,
    citecolor=citationmagenta,
    linkcolor=referenceblue,
    urlcolor=urlteal
}

\usepackage{booktabs}
\usepackage{amsfonts}
\usepackage{nicefrac}
\usepackage{microtype}
\usepackage[table]{xcolor}
\usepackage{amsmath}
\usepackage{mathtools}
\usepackage{tabularx}
\usepackage{amssymb}

\usepackage{graphicx}
\usepackage{float}
\usepackage{placeins}
\usepackage{subcaption}
\usepackage{stfloats}
\usepackage{capt-of}
\usepackage{siunitx}

\usepackage{glossaries-extra}

\setabbreviationstyle[acronym]{long-short}
\let\acrshort\glsxtrshort
\let\acrlong\glsxtrlong

\newacronym{ardm}{ARDM}{autoregressive diffusion model}
\newacronym{cada}{CADA}{Control-Augmented Data Assimilation}
\newacronym{da}{DA}{data assimilation}
\newacronym{pde}{PDE}{partial differential equation}
\newacronym{rmse}{RMSE}{root mean square error}
\newacronym{gt}{GT}{ground truth}
\newacronym{ttoda}{TTO-DA}{Test-Time Optimization based Data Assimilation}
\newacronym{tv}{TV}{Total Variation}
\newacronym{ddim}{DDIM}{Denoising Diffusion Implicit Model}
\newacronym{ks}{KS}{Kuramoto--Sivashinsky}
\newacronym{hct}{HCT}{high correlation time}
\newacronym{enkf}{EnKF}{Ensemble Kalman Filter}
\newacronym{bon}{BoN}{Best-of-$n$}
\newacronym{kl}{KL}{Kullback--Leibler}
\newacronym{snr}{SNR}{Signal-to-Noise Ratio}
\newacronym{film}{FiLM}{Feature-wise Linear Modulation}
\newacronym{4dvar}{4D-Var}{four-dimensional variational assimilation}
\newacronym{3dvar}{3D-Var}{three-dimensional variational assimilation}
\newacronym{era5}{ERA5}{ECMWF Reanalysis v5}
\newacronym{ds}{DS}{spatial downsampling}
\newacronym{ms}{MS}{strided masking}
\glsdisablehyper
\begin{document}

\maketitle

\begin{abstract}
  Autoregressive diffusion models (ARDMs) offer expressive priors for forecasting chaotic dynamics. However, maintaining consistency with sparse observations over long rollouts can require substantial test-time computation. We introduce Control-Augmented Data Assimilation (CADA), which trains a lightweight controller on top of a frozen ARDM to steer predictions toward future observations. Using a KL-regularized stochastic optimal control objective, the controller learns from short autoregressive rollouts to anticipate observations while staying close to the pretrained dynamics. At inference, it injects corrections at each denoising step without per-forecast optimization. Across two chaotic PDE benchmarks, CADA improves long-horizon accuracy, stability, and physical fidelity under sparse observations. On compact ERA5, it outperforms diffusion-based baselines in most regimes with over $10\times$ faster inference than conditional diffusion baselines. Applied to frozen GenCast, CADA also reduces temperature errors across 12--48-hour forecasts with spatially sparse observations. Code is available at \url{https://github.com/mandt-lab/cada}.
\end{abstract}

\section{Introduction}
\Glspl{ardm} are expressive models for high-dimensional spatiotemporal dynamics \citep{yang2023diffusion, yu2024language, huang2025selfforcing}, including scientific and geophysical applications \citep{pathak2024kilometerscaleconvectionallowingmodel, Mardani2025, srivastava2024precipitation, pandey2025heavytailed}. However, their reliance on one-step training with teacher forcing \citep{Williams1989} leaves them prone to error accumulation over long rollouts, especially in chaotic systems where small deviations rapidly amplify \citep{kalnay_2002_atmospheric, carrassi2018data, evensen2022data}. Observations from sensors, weather stations, or satellites can correct these errors, yet incorporating sparse, intermittent measurements into a pretrained \gls{ardm} is challenging. This raises the question: \emph{How can we steer long autoregressive trajectories toward observations without expensive test-time optimization?}

\begin{figure}[!tbp]
\centering

  \newlength{\panelheight}
  \setlength{\panelheight}{0.16\textheight} %

  \begin{subfigure}[b]{0.49\textwidth}
    \centering
    \begin{minipage}[c][\panelheight][c]{\linewidth}
      \centering
      \includegraphics[width=\linewidth]{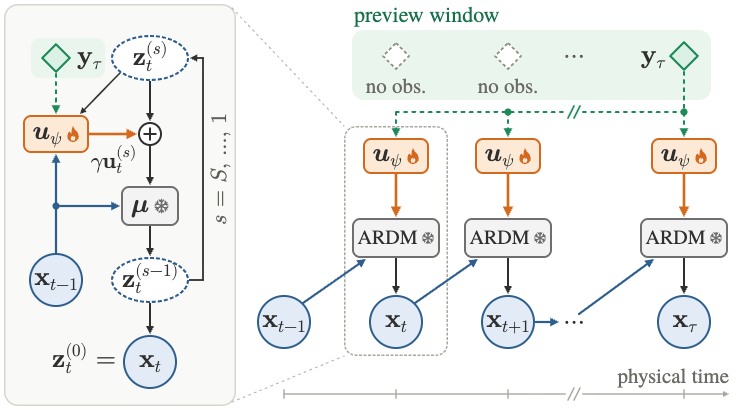}
    \end{minipage}
    \subcaption{Overview}
    \label{fig:overview}
  \end{subfigure}
  \hfill
  \begin{subfigure}[b]{0.49\textwidth}
    \centering
    \begin{minipage}[c][\panelheight][c]{\linewidth}
      \centering
      \includegraphics[width=\linewidth]{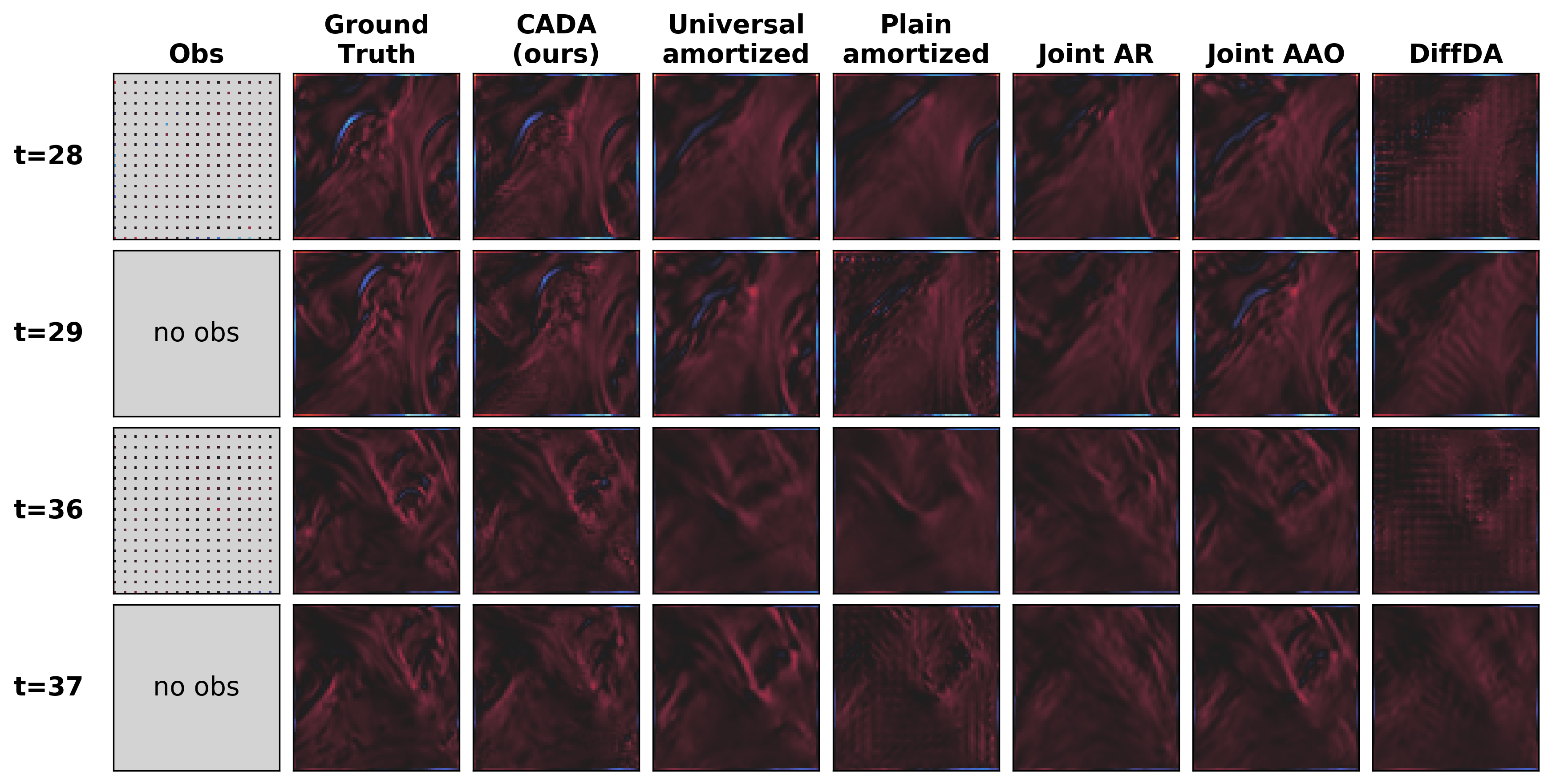}
    \end{minipage}
    \subcaption{\acrshort{era5}}
    \label{fig:era5vort}
  \end{subfigure}
  
  \caption{
  \textbf{A learned controller steers a frozen autoregressive diffusion model toward upcoming observations.}
(a) At each physical step, the controller $\vu_\psi$ (orange, trained) shifts every denoising step of the frozen \acrshort{ardm} (gray) using the previous state $\rvx_{t-1}$ and the (possibly empty) preview window (green). (b) \acrshort{era5} 500 hPa vorticity under \textsc{\acrshort{ms}}-4 observations; rows are forecast times, columns show observations (gray: none; dots: sparse measurements), ground truth, and reconstructions. \acrshort{cada} stays closest to ground truth, including at unobserved steps (full comparison: Fig.~\ref{fig:era5temp}).}
  
\end{figure}

We study this problem through \gls{da}, where forecasts of evolving physical systems are constrained by noisy, incomplete measurements \citep{kalnay_2002_atmospheric, carrassi2018data, evensen2022data, sanzalonso2023inverse}. Classical \gls{da} approaches, including ensemble Kalman filters and variational methods \citep{le1986variational, courtier1998ecmwf, rabier20004dvar, tr2006accounting}, rely on approximations or optimization procedures that can be challenging to combine with stochastic, non-Gaussian \gls{ardm} priors. Because observations arrive intermittently, effective corrections must also influence unobserved steps preceding a measurement. Diffusion models offer expressive priors that can be adapted through observation-conditioned training \citep{huang2024diffda} or inference-time guidance \citep{uehara2025inferencetimealignmentdiffusionmodels, rozet2023score, qu2024deep, manshausen2025generative}. However, guidance is particularly expensive for \glspl{ardm}: correcting a prediction using an observation several steps ahead can require differentiating through an autoregressive rollout. Repeating this for every assimilation window incurs substantial test-time cost, motivating an approach that learns the corrections in advance.

\paragraph{Method Overview.}
We introduce \gls{cada}, an amortized control framework that augments a frozen \gls{ardm} with a lightweight, observation-aware controller (Fig.~\ref{fig:overview}). We formulate \gls{da} as a \acrshort{kl}-regularized stochastic optimal control problem, balancing agreement with observations against deviations from pretrained dynamics. Instead of optimizing controls separately for each forecast, we train the controller offline through short autoregressive rollouts. At each physical step, it uses the noisy state, previous forecast, and a previewed observation to predict corrections injected into denoising steps. By learning how present corrections influence future observation consistency, \gls{cada} enables anticipatory guidance without changing \gls{ardm} parameters. At inference, assimilation reduces to a feed-forward rollout without per-instance optimization.

We evaluate \gls{cada} on two chaotic \glspl{pde}, Kolmogorov flow and Kuramoto--Sivashinsky, and a compact \acrshort{era5} weather benchmark \citep{hersbach2020era5}. Across sparse and delayed observations, \gls{cada} improves long-horizon accuracy, temporal stability, and physical fidelity over diffusion-based baselines in most settings. On compact \acrshort{era5}, it achieves over $10\times$ faster inference than conditional diffusion baselines. We also apply \gls{cada} to pretrained GenCast \citep{price2025gencast}, reducing full-field temperature errors over 12--48-hour forecasts with sparse spatial observations.

Our main contributions are:
\begin{itemize}
\item \textbf{Control-based guidance for pretrained \glspl{ardm}.} We formulate observation-conditioned autoregressive generation as \acrshort{kl}-regularized stochastic control, keeping the prior frozen.
\item \textbf{Amortized, anticipatory corrections.} We learn a reusable controller on short windows, injecting observation-aware corrections at denoising steps without per-forecast optimization.
\item \textbf{Accurate and efficient data assimilation.} We show improved long-horizon stability, physical fidelity, and efficiency on chaotic \glspl{pde} and weather benchmarks, including GenCast.
\end{itemize}

\section{Control Augmented Data Assimilation}
\label{sec:method}
We now formalize the \gls{da} setting and develop the mathematical foundations underlying our approach.

\subsection{Problem Statement: Chaotic Forecasting with Sparse Observations}
\label{subsec:problem}
Autoregressive forecasts of physical systems are vulnerable to error accumulation, especially under chaotic dynamics, where small deviations can rapidly amplify. In practice, such forecasts are stabilized by sequentially incorporating partial observations, a process known as \gls{da}. Our goal is to guide a pretrained autoregressive model toward forecasts that remain consistent with incoming sparse observations while reducing long-horizon drift.

\begin{figure}[t]

  \begin{subfigure}[t]{0.534\textwidth}
    \centering
    \includegraphics[width=\linewidth]{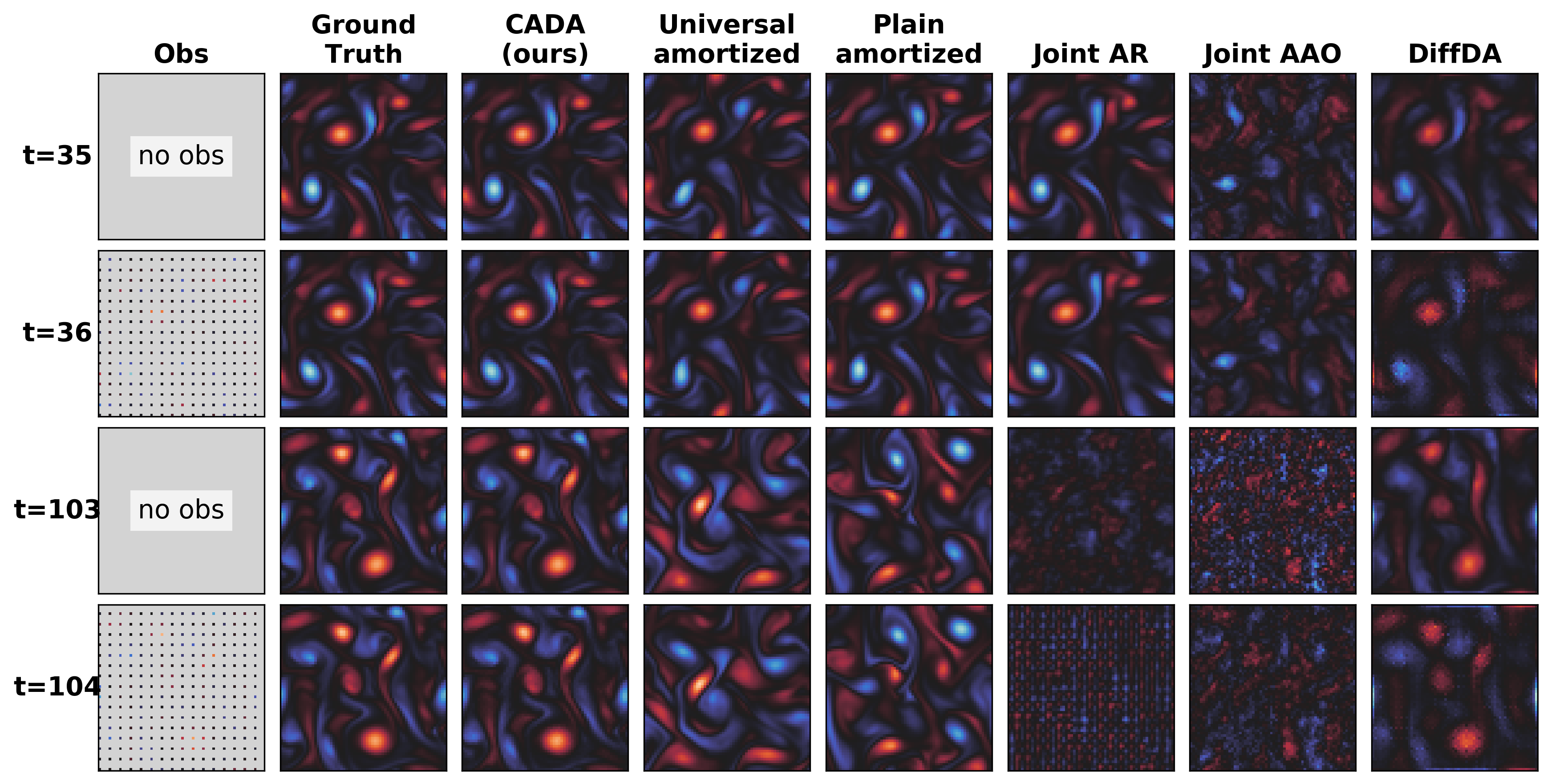}
    \subcaption{Kolmogorov Flow}
    \label{fig:compare}
  \end{subfigure}
  \hfill
  \begin{subfigure}[t]{0.445\textwidth}
    \centering
    \includegraphics[width=\linewidth]{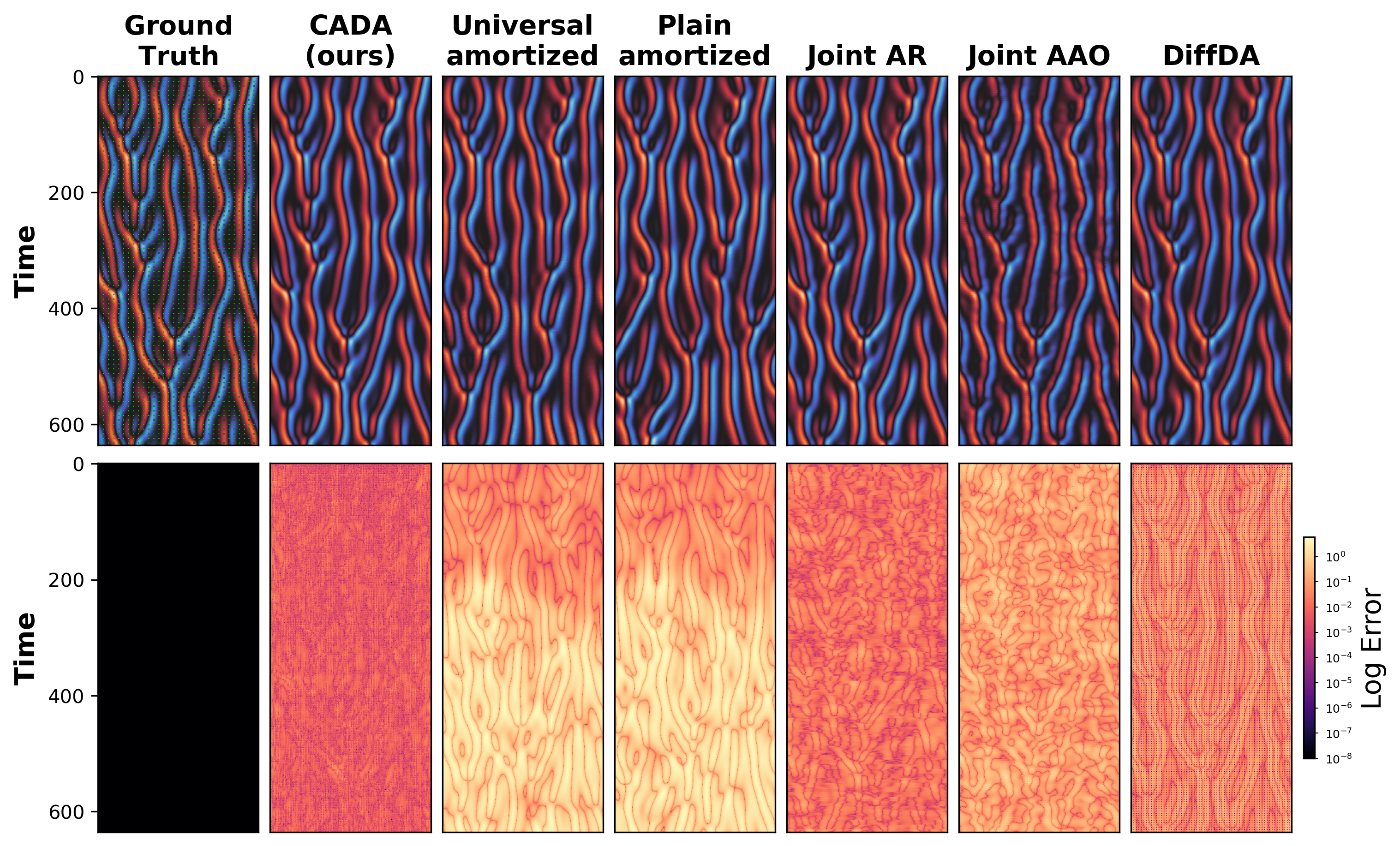}
    \subcaption{\gls{ks}}
    \label{fig:ks}
  \end{subfigure}
  \caption{\textbf{\acrshort{cada} stays accurate over long rollouts with sparse \acrshort{ms}-4 observations.} (a) Kolmogorov vorticity at selected steps of a 180-step rollout (rows); columns show observations (gray: none; dots: sparse measurements), ground truth, and reconstructions. (b) \acrshort{ks} space--time fields over 640 steps (vertical: time; horizontal: space); top: fields, with observations in green on the ground truth; bottom: log pointwise error (darker is lower). \acrshort{cada} preserves coherent structures with the least error.}
  \label{fig:pde_summary}
\end{figure}

We consider system states $\rvx_t$ at physical times $t \in \mathbb{N}$. Observations are available only at times $\mathcal{T}\subseteq\mathbb{N}$, denoted $\rvy_{\gT}\triangleq\{\rvy_\tau\}_{\tau\in\gT}$. In weather forecasting, for example, they may come from stations, radars, or satellites. We assume observations are noisy and partial, hence insufficient to recover the full state $\rvx_t$. We are also given a pretrained autoregressive forecasting model
\begin{equation}
\label{eq:base-path}
\gQ
\;=\;
q_0(\rvx_0)\;\prod_{t\ge 1}
q\big(\rvx_{t}\,\big|\,\rvx_{t-1}\big).
\end{equation}
Here $q_0$ is the initial-state distribution; we later instantiate the transition $q$ as a conditional diffusion model \citep{ho2020denoising,pandey2023complete} whose parameters stay fixed and are left implicit (Sec.~\ref{sec:guided-ardm}). The objective is to incorporate $\rvy_{\gT}$ into the forecast while keeping the pretrained prior $\gQ$ fixed, i.e., without retraining it.

We need a cost function $\Phi(\rvx_\tau; \rvy_\tau)$ that measures the compatibility between a forecast state $\rvx_\tau$ and an observation $\rvy_\tau$ to optimally assimilate observations. For example, in a common observation model $\rvy_\tau = A(\rvx_\tau)$ (where $A$ is some lossy data degradation operator such as blurring, masking, or spatial downsampling), one may take $\Phi = \|\rvy_\tau - A(\rvx_\tau)\|_2^2$ (more details on operators can be found in App.~\ref{app:obs-ops}). Equivalently, this cost induces a likelihood model $p(\rvy_\tau |\rvx_\tau) \propto \exp({-\frac{1}{\beta} \Phi(\rvx_\tau; \rvy_\tau)})$, where $\beta>0$ is a temperature parameter that controls tolerance to observation mismatch.

A principled way to assimilate observations is Bayesian inference. To this end, the optimal sampling distribution is the \emph{tilted posterior} distribution that balances prior dynamics with observations:
\begin{equation}
    \label{eq:tilt_distribution}
    \gP^* \;\propto\; \gQ\,\exp\!\biggl(-\frac{1}{\beta}\sum_{\tau\in\gT} \Phi(\rvx_\tau; \rvy_\tau)\biggr).
\end{equation}

In most cases of interest, this posterior distribution is intractable; however, notable exceptions exist in the literature. For example, classical \gls{da} methods such as Kalman filtering oftentimes assume that $\mathcal{Q}$ is a Gauss-Markov model and $A(\cdot)$ is linear~\citep{sanzalonso2023inverse}, in which case the Bayesian updates are tractable. However, such assumptions severely limit the expressivity (e.g., multimodality) of $\mathcal{Q}$. Interestingly, $\gP^*$ is the optimizer of the following variational problem,
\begin{equation}
\label{eq:arrival-cost}
\gC(\gP) \;\triangleq\; \sum_{\tau\in\gT} \E_{\rvx_\tau\sim \gP}\big[\Phi(\rvx_\tau;\,\rvy_\tau)\big] + \beta\,\KL\!\big(\gP \,\|\, \gQ\big)
\end{equation}
(proof in App.~\ref{app:proof-tilt}). This is a stochastic optimal control objective: a controlled process pays a $\beta$-weighted \gls{kl} cost for deviating from the uncontrolled dynamics $\gQ$, the view that also underlies control-based guidance and fine-tuning of diffusion models~\citep{pandey2025variational,domingo-enrich2025adjoint}. Restricting $\gP$ to a tractable family yields a variational inference problem~\citep{zhang2018advances}. While one may sample directly from the unnormalized target in Eq.~\ref{eq:tilt_distribution}~\citep{vargas2023denoising,richter2024improved}, we instead optimize Eq.~\ref{eq:arrival-cost} by learning guided autoregressive dynamics. We next define the unguided and guided dynamics (Sec.~\ref{sec:guided-ardm}), how the guidance is learned (Sec.~\ref{sec:learning-controls}), and practical design choices (Sec.~\ref{sec:practical_choices}).

\subsection{Diffusion-based Dynamics Modeling and Control}
\label{sec:guided-ardm}

We use conditional \glspl{ardm}~\citep{yang2023diffusion,ruhling2023dyffusion,price2025gencast} and distinguish physical time indices $t\in\mathbb{N}$ (in subscripts) from diffusion denoising sub-steps $s\in \{S,\dots,1\}$ (in superscripts).

As in Sec.~\ref{subsec:problem}, let
$\gQ = q_0(\rvx_0)\prod_{t\ge 1} q(\rvx_t\mid \rvx_{t-1})$
denote the pretrained autoregressive prior. Each transition $q(\rvx_t\mid \rvx_{t-1})$ is defined as the marginal endpoint distribution of a conditional diffusion sampler:
\begin{equation}
\label{eq:unguided_ardm}
q(\rvx_{t}\!\mid \rvx_{t-1})
=
\int \!\left[
\prod_{s=1}^{S}
q\left(\rvz_{t}^{(s-1)} \,\big|\, \rvz_{t}^{(s)};\, \rvx_{t-1}\right)
\right]\,
q\left(\rvz_{t}^{(S)}\right)\;
d \rvz_{t}^{(1:S)} .
\end{equation}
The denoising transitions are Gaussian, $q(\rvz_{t}^{(s-1)}\mid \rvz_{t}^{(s)}; \rvx_{t-1})
=
\gN\!(
\vmu(\rvz_{t}^{(s)}, \rvx_{t-1}, s),
\sigma_s^2 \mI_d
),$ where $\vmu$ is the pretrained denoiser, $\sigma_s$ is the standard deviation of sub-step $s$, $d$ is the state dimension, $\rvz_t^{(s)}$ are intermediate noisy states, and $q(\rvz_t^{(S)})=\gN(\mathbf{0},\mI_d)$ is the initial noise prior. Thus, each autoregressive step samples $\rvx_t \equiv \rvz_t^{(0)}$ by running $S$ denoising steps conditioned on $\rvx_{t-1}$. We refer to this process as the \emph{unguided \gls{ardm} dynamics}.

As in Sec.~\ref{subsec:problem}, our goal is to steer the pretrained unguided process $\gQ$ toward trajectories consistent with upcoming observations while remaining close to the original dynamics. We model the variational distribution $\gP$ as an autoregressive process augmented with control variables $\mU_t$:
\begin{equation}
\label{eq:controlled-path}
\gP
\;=\;
p_0(\rvx_0)\;\prod_{t\ge 1}
p\big(\rvx_{t}\,\big|\,\rvx_{t-1}; \mU_{t}\big).
\end{equation}
Here $p_0=q_0$, since controls act only on transitions. Analogously to $\gQ$, each controlled transition is defined as
\begin{equation}
\label{eq:controlled-kernel}
p(\rvx_{t}\mid \rvx_{t-1};\mU_{t})
=
\int
\left[
\prod_{s=1}^{S}
p\!\left(\rvz_{t}^{(s-1)} \!\mid\! \rvz_{t}^{(s)};\rvu_{t}^{(s)},\rvx_{t-1}\right)
\right]
q\left(\rvz_{t}^{(S)}\right)
\mathrm{d}\rvz_{t}^{(1:S)},
\end{equation}
where $\mU_t=(\rvu_t^{(1)},\ldots,\rvu_t^{(S)})$ denotes the controls applied across denoising sub-steps. At each denoising sub-step, the control $\rvu_t^{(s)}$ perturbs the noisy input to the pretrained denoiser, yielding
\begin{equation}
\label{eq:controlled-substep}
p\!\left(\rvz_{t}^{(s-1)} \!\mid\! \rvz_{t}^{(s)};\rvu_{t}^{(s)},\rvx_{t-1}\right)
\triangleq
\gN\!\left(
\vmu\!\left(\rvz_{t}^{(s)} + \gamma \rvu_{t}^{(s)},\,\rvx_{t-1},\,s\right),
\sigma_{s}^2 \mI_d
\right),
\end{equation}
where $\vmu$ is the fixed pretrained denoiser and $\gamma>0$ controls the strength of the input shift. Thus, $\gP$ preserves the pretrained denoising model while introducing additive controls in the noisy state space. Setting $\rvu_t^{(s)}\equiv\mathbf{0}$ recovers the unguided kernel, so $\gQ$ is the special case of $\gP$ without control.

Given controls $\mU_t$, we can roll out the controlled autoregressive process, evaluate the observation costs along the trajectory, and differentiate through the rollout to optimize the variational objective in Eq.~\ref{eq:arrival-cost}. We refer to this parameterization as a \emph{guided \gls{ardm}}. We next describe how the controls are learned from upcoming observations over a fixed preview horizon.

\subsection{Learning the controls}
\label{sec:learning-controls}

We learn the controls $\mU_{t}$ by optimizing the variational objective in Eq.~\ref{eq:arrival-cost}. A direct alternative is to treat $\{\rvu_t^{(s)}\}$ as free variables and optimize them separately for each assimilation window at test time, analogous to test-time scaling for diffusion models~\citep{uehara2025inferencetimealignmentdiffusionmodels}. We refer to this ablation as \emph{\gls{ttoda}}. Another inference-time alternative is \emph{reconstruction guidance}, which adjusts each diffusion step using gradients of the observation cost, e.g., $\nabla_{\rvx}\Phi(\rvx;\rvy)$, without a learned controller. However, without a direct estimate $\E[\rvx_{\tau}\mid\rvz_t^{(s)}]$ of the future state at an observation index $\tau > t$, both approaches require backpropagation through the \gls{ardm} rollout for each new forecast, making them prohibitively expensive for long-horizon assimilation.

We instead amortize the controls with a controller trained offline. During training, we roll out the guided \gls{ardm} in Eq.~\ref{eq:controlled-path} and evaluate the observation costs in Eq.~\ref{eq:arrival-cost} at observation times $\tau$. Since each controlled transition requires specifying $\mU_t$, we parameterize the control at physical time $t$ and denoising sub-step $s$ as
\begin{equation}
\label{eq:control-policy}
\rvu_{t}^{(s)} = \vu_\psi\!\big(\rvz_{t}^{(s)},\;\rvx_{t-1},\;\bar{\rvy}_t,\;s\big),
\end{equation}
where $\rvz_t^{(s)}$ is the current noisy state, $\rvx_{t-1}$ is the previous forecast, and $\bar{\rvy}_t=(\rvy_{\tau_t},\mM_{\tau_t},\Delta_{t,\tau_t})$ is the \emph{preview}: the nearest observation $\tau_t\ge t$ in the current window, with its mask and lead time (null if none; App.~\ref{app:selector}). We optimize only the controller parameters $\psi$, keeping the pretrained \gls{ardm} fixed. The amortized controller produces corrections with a single forward pass at test time. We call the full framework \emph{\gls{cada}}.

\newsavebox{\algbox}

\begin{figure}[t]
\centering

\sbox{\algbox}{%
\begin{minipage}[t]{0.60\textwidth}
\vspace{0pt}
\begin{algorithm}[H]
\caption{Controlled Transition ($\rvx_{t-1} \to \rvx_t$)}
\label{alg:core_transition}
\DontPrintSemicolon
\footnotesize
\SetAlgoNlRelativeSize{-1}
\SetInd{0.35em}{0.75em}

\KwIn{$\rvx_{t-1}$; future stream $\{\rvy_j\}_{j\in\mathcal T}$; horizon $\Lambda$; control scale $\gamma$}
\KwOut{Next state $\rvx_t$}

\textbf{Context:} Build preview $\bar{\rvy}_{t}$ from $\{\rvy_j\}_{j\in\mathcal T}$ (App.~\ref{app:selector})\;
\textbf{Init:} Sample $\rvz_t^{(S)} \sim q(\rvz_t^{(S)})$\;

\For{$s = S,\dots,1$}{
  \begingroup\setlength{\fboxsep}{0pt}\colorbox{orange!12}{$\rvu_t^{(s)} \leftarrow \vu_\psi(\rvz_t^{(s)},\rvx_{t-1},\bar{\rvy}_t,s)$}\endgroup \tcp*{Eq.~\ref{eq:control-policy}}
  \begingroup\setlength{\fboxsep}{0pt}\colorbox{orange!12}{$\tilde{\rvz} \leftarrow \rvz_t^{(s)} + \gamma \rvu_t^{(s)}$}\endgroup \tcp*{Eq.~\ref{eq:controlled-substep}}
  $\rvz_t^{(s-1)} \leftarrow \textsc{DenoiseStep}(\tilde{\rvz},\rvx_{t-1},s)$ \tcp*{\acrshort{ddim}}
}

\Return{$\rvx_t \leftarrow \rvz_t^{(0)}$}\;
\end{algorithm}
\vspace{2pt}
\colorbox{orange!12}{Highlighted} lines 4--5 are the \acrshort{cada} additions: one controller call per sub-step predicts and injects the control; line~6 is the frozen \acrshort{ardm}'s \acrshort{ddim} step. Full notation: Alg.~\ref{alg:controlled-step}.
\end{minipage}%
}

\makebox[\textwidth][c]{%
\usebox{\algbox}
\hspace{0.025\textwidth}
\begin{minipage}[t]{0.33\textwidth}
\vspace{0pt}
\centering
\captionsetup{justification=raggedright,singlelinecheck=false}
\captionof{table}{\textbf{\acrshort{cada} best preserves physical diagnostics.}
Time-avg.\ abs.\ error (Eq.~\ref{eq:physics-error}), \acrshort{ms}-8: dissipation rate (Kolmogorov, 60 steps), total variation (\acrshort{ks}, 640 steps).}
\label{tab:physics_diagnostics}

\footnotesize
\setlength{\tabcolsep}{3pt}
\begin{tabular}{@{}lcc@{}}
\toprule
\textbf{Method} & \textbf{Diss.} $\downarrow$ & \textbf{\acrshort{tv}} $\downarrow$ \\
\midrule
\textbf{\acrshort{cada}} & \textbf{2.8e-4} & \textbf{0.12} \\
Joint AAO                & 9.2e-3 & 3.98 \\
Joint AR                 & 9.0e-4 & 1.56 \\
Plain amortized          & 5.7e-4 & 1.85 \\
Universal amortized      & 8.5e-4 & 1.91 \\
DiffDA                   & 9.1e-3 & 2.70 \\
\acrshort{ttoda}         & 8.4e-3 & 3.12 \\
\acrshort{bon}           & 1.0e-2 & 4.02 \\
\bottomrule
\end{tabular}
\end{minipage}%
}

\end{figure}

\gls{cada} learns a reusable controller from short \emph{preview windows}, so trajectory-level optimization is performed once offline rather than per forecast. At inference, assimilation reduces to rolling out the pretrained \gls{ardm} with feed-forward control corrections. Empirically, this amortized guidance substantially reduces inference cost while producing more stable and physically faithful \gls{da} trajectories than reconstruction guidance and \gls{ttoda} baselines (see Sec.~\ref{sec:exp}).

\subsection{Training and Inference}
\label{sec:practical_choices}

In both training and inference, we assume access to the observation schedule $\mathcal{T}$ within a preview window of length $\Lambda$, together with the corresponding observations $\{\rvy_j\}_{j\in\mathcal{T}}$.

\paragraph{Forward pass through the preview.}
Given a preview window of length $\Lambda$, we sample an initial state and roll out the controlled dynamics in Eq.~\ref{eq:controlled-kernel}. At each physical step $t$ in the window, we build the preview $\bar{\rvy}_t$, and the controller produces per-step denoising controls
$\rvu_t^{(s)}=\vu_\psi(\rvz_t^{(s)},\rvx_{t-1},\bar{\rvy}_t,s)$
as in Eq.~\ref{eq:control-policy}, which are injected as noisy-state shifts in the reverse diffusion transitions in Eq.~\ref{eq:controlled-substep}; Alg.~\ref{alg:core_transition} summarizes the resulting transition $\rvx_{t-1}\!\to\!\rvx_t$.

\paragraph{Training objective restricted to the preview.}
During training, we evaluate observation costs only at timesteps within the window that contain observations. The controller is optimized using only the information available in its preview context $\bar{\rvy}_t$. Let
$\gP_\psi(\rvx_{t_0+1:t_0+\Lambda}\mid \rvx_{t_0})$
denote the path distribution over a window of $\Lambda$ steps induced by the controlled kernel with controls from $\vu_\psi$, and let $\mathcal{T}^\Lambda=\mathcal{T}\cap[t_0{+}1,\,t_0{+}\Lambda]$ be the observation times in the window. We minimize the windowed variational objective $
\mathcal{L}(\psi)
=
\sum_{\tau\in\mathcal{T}^\Lambda}
\E_{\gP_\psi}
\big[\Phi(\rvx_\tau;\rvy_\tau)\big]
+
\beta\,
\KL\!\big(
\gP_\psi(\cdot\mid \rvx_{t_0})
\,\|\,
\gQ(\cdot\mid \rvx_{t_0})
\big),$
where $\gQ(\cdot\mid\rvx_{t_0})$ is the corresponding unguided \gls{ardm} rollout. We approximate expectations with Monte Carlo rollouts, average the observation cost over the arrivals in $\mathcal{T}^\Lambda$ and also evaluate it on Tweedie estimates at intermediate denoising sub-steps, optimize $\psi$ by stochastic gradient descent, and choose $\beta$ by grid search. Since the marginal one-step \gls{kl} is intractable, we use a tractable surrogate matched to our parameterization; see App.~\ref{app:kl_surrogate}. Alg.~\ref{alg:training} in App.~\ref{app:algo} gives the full training loop.

\paragraph{Inference by chunked re-anchoring.}
At test time, we apply the learned controller with a single forward pass per denoising sub-step, requiring no test-time optimization. We advance autoregressively and match the training setup by re-anchoring every $\Lambda$ steps: the forecast horizon is partitioned into chunks of length $\Lambda$, and each preview context $\bar{\rvy}_t$ uses only observations scheduled within the current chunk. The final state of one chunk initializes the next. Alg.~\ref{alg:sampling-chunked} in App.~\ref{app:algo} specifies the full chunking and handoff procedure. We implement $\textsc{DenoiseStep}$ with \acrshort{ddim}~\citep{song2021ddim}, but the preview-and-control mechanism is sampler-agnostic and only requires access to diffusion sub-steps.

\section{Related Work}
\paragraph{Guidance in Diffusion Models.} Some existing works on guidance approximate the noisy likelihood score by estimating the diffusion posterior $q(\rvz^{(0)}\mid\rvz^{(s)})$ \citep{chung2022diffusion, song2023pseudoinverseguided, kawar2022denoising, pandey2024fast, pokle2024trainingfree}. While this can result in accurate guidance and faster sampling, a large proportion of these methods are limited to linear inverse problems, which further limits their application. More recent works \citep{pandey2025variational, rout2025rbmodulation} alleviate some of these problems by formulating guidance as optimal control. Our method amortizes the controls in a separate training stage in the framework of autoregressive diffusion models. There has also been recent work in fine-tuning diffusion models \citep{domingo-enrich2025adjoint, clark2024directly, fan2023reinforcement, pandey2026hierarchicalvariationalpoliciesrewardguided} which is complementary to our proposed framework.

\paragraph{Data assimilation.} Several recent works utilize diffusion models for \gls{da}. \citet{rozet2023score} train score-based diffusion models on short trajectory segments to generate full long trajectories during inference. Their framework has been also applied in \citet{qu2024deep} and \citet{manshausen2025generative}. However, these approaches rely on inference-time-only guidance, lacking mechanisms that avoid error accumulation during observational gaps. To mitigate this, established practices utilize larger temporal windows of observations, as seen in classical 4D-Var \citep{le1986variational} and iterative smoothers \citep{bocquet_2014_iterative}, which provide the multi-step constraints necessary to better constrain the physical manifold \citep{nie_2023_influence}. Autoregressive methods \citep{huang2024diffda, shysheya2024conditional, gao2024bayesian} improve stability but remain computationally expensive due to inference-time optimization. Latent space-based approaches  \citep{foroumandi2025harnessing, fan2026physically} reduce dimensionality but introduce reconstruction biases and latent-physical decoupling errors. We address these limitations by integrating learned feedback controls into autoregressive diffusion denoising, enabling inference as a single forward rollout with long-horizon stability and computational efficiency.

\begin{table*}[t]
\centering
\caption{
\textbf{\acrshort{cada} attains the lowest \acrshort{rmse} in most regimes on Kolmogorov flow and Kuramoto--Sivashinsky (\acrshort{ks}).} Columns: short/long rollouts (Kolmogorov 60/180, \acrshort{ks} 140/640 steps) under \acrshort{ds}-$k$ and \acrshort{ms}-$k$ observations (Sec.~\ref{par:obs-regimes}). Rows: baselines and ablations (\acrshort{ttoda}: no amortization; \acrshort{bon}: best of 16 unguided samples). Best in bold. \acrshort{hct}, \acrshort{era5}, classical baselines: Tabs.~\ref{tab:hct_clean}, \ref{tab:rmse-mr4}, \ref{tab:enkf}.
}
\label{tab:rmsd_clean}
\setlength{\tabcolsep}{1.5pt}
\begin{scriptsize}
\begin{tabular*}{\linewidth}{@{\extracolsep{\fill}}l *{16}{c}}
\toprule
& \multicolumn{2}{c}{\textbf{DS-2}} & \multicolumn{2}{c}{\textbf{DS-4}} &
  \multicolumn{2}{c}{\textbf{DS-8}} & \multicolumn{2}{c}{\textbf{DS-16}} &
  \multicolumn{2}{c}{\textbf{MS-2}} & \multicolumn{2}{c}{\textbf{MS-4}} &
  \multicolumn{2}{c}{\textbf{MS-8}} & \multicolumn{2}{c}{\textbf{MS-16}} \\
\cmidrule(lr){2-3}\cmidrule(lr){4-5}\cmidrule(lr){6-7}\cmidrule(lr){8-9}
\cmidrule(lr){10-11}\cmidrule(lr){12-13}\cmidrule(lr){14-15}\cmidrule(lr){16-17}
& {short} & {long} & {short} & {long} & {short} & {long} & {short} & {long} &
{short} & {long} & {short} & {long} & {short} & {long} & {short} & {long} \\
\midrule
\multicolumn{17}{c}{\textbf{Kolmogorov flow}} \\
\midrule
\textbf{\acrshort{cada} (ours)} & \bfseries 0.016 & \bfseries 0.016 & \bfseries 0.020 & \bfseries 0.020 & \bfseries 0.138 & 0.351 & 0.428 & 0.638 & \bfseries 0.017 & \bfseries 0.017 & \bfseries 0.024 & \bfseries 0.024 & \bfseries 0.060 & \bfseries 0.286 & 0.424 & 0.641 \\
Joint AAO & 0.041 & 0.045 & 0.210 & 0.189 & 0.380 & 0.244 & 0.419 & 0.249 & 0.141 & 0.171 & 0.358 & 0.559 & 0.465 & 0.523 & 0.499 & 0.433 \\
Joint AR & 0.038 & 0.031 & 0.185 & 0.115 & 0.366 & \bfseries 0.218 & \bfseries 0.413 & \bfseries 0.245 & 0.046 & 0.129 & 0.152 & 0.261 & 0.404 & 0.574 & 0.448 & 0.436 \\
Plain amortized & 0.109 & 0.814 & 0.229 & 1.033 & 0.712 & 1.276 & 0.799 & 1.175 & 0.245 & 0.454 & 0.302 & 0.477 & 0.316 & 0.479 & \bfseries 0.309 & 0.471 \\
Universal amortized & 0.295 & 1.398 & 1.061 & 1.566 & 1.612 & 1.766 & 1.785 & 1.862 & 0.186 & 0.397 & 0.323 & 0.469 & 0.351 & 0.483 & 0.353 & 0.489 \\
DiffDA & \NA & \NA & \NA & \NA & \NA & \NA & \NA & \NA & 0.230 & 0.232 & 0.241 & 0.283 & 0.358 & 0.364 & 0.386 & \bfseries 0.391 \\
\arrayrulecolor{lightgray}\hline\arrayrulecolor{black}
\acrshort{ttoda} & 0.040 & 0.243 & 0.027 & 0.115 & 0.156 & 0.401 & 0.491 & 1.280 & 0.078 & 0.298 & 0.113 & 0.357 & 0.215 & 0.433 & 0.526 & 0.611 \\
\acrshort{bon} & 0.258 & 0.420 & 0.264 & 0.440 & 0.299 & 0.442 & 1.759 & 1.992 & 0.265 & 0.447 & 0.266 & 0.454 & 0.306 & 0.488 & 0.846 & 2.089 \\
\midrule
\multicolumn{17}{c}{\textbf{\acrlong{ks}}} \\
\midrule
\textbf{\acrshort{cada} (ours)} & \bfseries 0.006 & \bfseries 0.006 & \bfseries 0.006 & \bfseries 0.006 & \bfseries 0.009 & \bfseries 0.009 & \bfseries 0.019 & \bfseries 0.319 & \bfseries 0.006 & \bfseries 0.006 & \bfseries 0.011 & \bfseries 0.011 & \bfseries 0.011 & \bfseries 0.096 & \bfseries 0.018 & \bfseries 0.309 \\
Joint AAO & 0.017 & 0.017 & 0.091 & 0.092 & 0.417 & 0.424 & 0.922 & 0.939 & 0.045 & 0.038 & 0.210 & 0.195 & 0.614 & 0.599 & 1.228 & 1.208 \\
Joint AR & 0.018 & 0.018 & 0.091 & 0.093 & 0.413 & 0.428 & 0.902 & 0.920 & 0.026 & 0.009 & 0.041 & 0.032 & 0.134 & 0.136 & 0.726 & 1.187 \\
Plain amortized & 0.041 & 9.787 & 0.146 & 10.73 & 1.859 & 11.81 & 5.979 & 12.830 & 0.034 & 1.163 & 0.036 & 1.165 & 0.039 & 1.211 & 0.034 & 1.189 \\
Universal amortized & 0.043 & 5.574 & 0.146 & 6.210 & 2.096 & 6.947 & 5.032 & 7.408 & 0.041 & 1.098 & 0.044 & 1.197 & 0.048 & 1.239 & 0.046 & 1.278 \\
DiffDA & \NA & \NA & \NA & \NA & \NA & \NA & \NA & \NA & 0.046 & 0.099 & 0.132 & 0.133 & 0.251 & 0.324 & 0.482 & 0.499 \\
\arrayrulecolor{lightgray}\hline\arrayrulecolor{black}
\acrshort{ttoda} & 0.016 & 8.288 & 0.009 & 0.053 & 0.418 & 0.634 & 0.930 & 2.613 & 0.016 & 0.298 & 0.114 & 0.363 & 0.081 & 0.580 & 1.573 & 3.559 \\
\acrshort{bon} & 0.046 & 1.257 & 0.046 & 1.498 & 0.048 & 3.122 & 2.827 & 4.598 & 0.045 & 1.987 & 0.046 & 1.644 & 0.049 & 2.128 & 1.241 & 1.948 \\
\bottomrule
\end{tabular*}
\end{scriptsize}
\end{table*}
\section{Experiments}
\label{sec:exp}

We evaluate \gls{cada} on two chaotic \gls{pde} benchmarks, compact \gls{era5}, and larger-scale GenCast. Tests span six observation regimes, factor 16, nonlinear observations, and a one-step preview. We compare five diffusion-based baselines, two \gls{cada} ablations, and three classical methods. We ask: \emph{(i)} Does preview-aware control reduce drift under sparse and delayed observations? \emph{(ii)} Do gains persist over long chaotic rollouts? \emph{(iii)} Are improvements due to amortized control rather than test-time optimization or brute-force search?

\paragraph{Datasets.}
We consider two canonical chaotic \gls{pde} benchmarks: the 1D \gls{ks} equation and 2D Kolmogorov flow. Both exhibit nonlinear instabilities and long-range correlations, making them challenging testbeds for sparse-observation \gls{da} \citep{du2021evolutional,lippe2023pde}. We additionally evaluate on a compact \gls{era5} assimilation setup and in a larger-scale GenCast weather study. The governing \gls{pde}s and data-generation protocols are detailed in App.~\ref{app:expdet}.

\paragraph{Observation Regimes.}
\label{par:obs-regimes}
For each dataset, we evaluate six assimilation regimes designed to probe robustness to spatial sparsity and temporal delay, following settings inspired by \citet{rozet2023score} and \citet{shysheya2024conditional}. The first group, \emph{downsampled} observations \acrshort{ds}-$k$ with $k\in\{2,4,8\}$, observes the full field at every simulator step, but only as averages over non-overlapping $k{\times}k$ blocks (App.~\ref{app:obs-ops}). The second group, \emph{masked} observations \acrshort{ms}-$k$ with the same factors, observes the field only at every fourth simulator step and only at every $k$-th grid point along each spatial axis; all other steps and locations are unobserved. We extend both groups to factor 16, yielding a substantially tighter spatial information bottleneck without changing the temporal schedule. To test robustness beyond linear measurements, we also use the nonlinear masked operator $\rvy=(\mM\odot\rvx)^2+0.01(\mM\odot\mathbf{n})$, with $\mathbf{n}\sim\gN(\mathbf{0},\mI_d)$, at factors $\{2,4,8,16\}$. The operator is applied in normalized model space and uses the same masks and four-step cadence as the linear \acrshort{ms} setting. Squaring makes the observation map many-to-one by discarding sign information, while the additive noise prevents exact inversion even at observed locations. In all regimes, observation metadata, such as masks and operator parameters, are provided with each observation; see App.~\ref{app:obs-ops}.

\paragraph{Experimental Setup.}
For each observation regime, we train a separate \emph{controller} on top of a fixed pretrained \gls{ardm}, using one \gls{ardm} per dataset. Architectural details for the \gls{ardm} and controller are given in App.~\ref{app:ardm_details} and App.~\ref{app:ctrl-net}. \gls{cada} uses \gls{ddim} sampling with $S{=}3$ denoising sub-steps per \gls{ardm} transition. The preview window is $\Lambda{=}16$ for Kolmogorov flow and \gls{era5}, and $\Lambda{=}54$ for \gls{ks}; within each anchored window, observations are selected by the nearest-arrival rule (App.~\ref{app:selector}). The factor-16 and nonlinear experiments retain these preview lengths. We separately set $\Lambda{=}1$ for \acrshort{ms}-2/4/8 while preserving the original four-step observation cadence, operator, forecast horizon, and pretrained prior. Consequently, only transitions whose one-step preview contains a scheduled arrival receive an observation; the remaining transitions use a null observation context. This design changes the amount of future information available to the controller without conflating preview length with observation frequency. We report both short- and long-horizon rollouts to separate near-term correction from long-term stability: 60/180 steps for Kolmogorov flow, 140/640 steps for \gls{ks}, and 60 steps for \gls{era5}.

\paragraph{Evaluation Metrics.}
We evaluate \gls{da} using trajectory-level and physics-aware metrics. Forecast accuracy is measured by time-averaged \acrshort{rmse}, and temporal stability by \acrshort{hct}. To assess physical fidelity beyond pointwise error, we compute a domain-specific diagnostic $D$, \emph{total variation} (\acrshort{tv}, $\int|\partial_\xi u|\,d\xi$) for \gls{ks} and the \emph{dissipation rate} $\varepsilon$ for Kolmogorov flow (App.~\ref{app:expdet}), on every forecast state $\rvx_t$ and ground-truth state $\rvx^\star_t$ of an $H$-step rollout, and report the time-averaged absolute error
\begin{equation}
\label{eq:physics-error}
e_D \;=\; \frac{1}{H}\sum_{t=1}^{H}\big|D(\rvx_t)-D(\rvx^\star_t)\big|,
\end{equation}
averaged over test trajectories.

\paragraph{Baselines.}
We compare against five diffusion-based \gls{da} baselines. \textit{Joint AAO} \citep{rozet2023score} learns local joint scores over short segments under a $k$-order Markov factorization and samples trajectories \emph{all-at-once}; although principled, it is memory-intensive for long-horizon stability at the cost of additional neural evaluations. We also include the conditional-score variants of \citet{shysheya2024conditional}: \textit{Plain amortized}, trained with a fixed conditioning length $C$, and \textit{Universal amortized}, trained with $C\sim\mathrm{U}(\{0,\dots,2k\})$ (Markov order $k$ as above) and masking to support any split of a fixed input window into $C$ conditioning and $P$ predicted states. Both conditional variants can be combined with reconstruction guidance for partial observations. \textit{DiffDA} \citep{huang2024diffda} conditions each denoising transition on a forecast state and sparse observations. Because it enforces observations through inpainting-style masks over observed grid columns, we report it only for masked (\acrshort{ms}-$k$) regimes.

\paragraph{Classical baselines.}
On Kolmogorov flow, we additionally report \textit{\gls{enkf}}, \textit{\gls{3dvar}} \citep{courtier1998ecmwf}, and \textit{\gls{4dvar}} \citep{rabier20004dvar}. \gls{enkf} performs sequential ensemble updates using sample covariances; \gls{3dvar} solves a per-arrival maximum-a-posteriori problem; and \gls{4dvar} optimizes an assimilation window's initial state so its rollout matches all observations.

\paragraph{Ablations.}
(i) \textit{\gls{ttoda}} performs non-amortized test-time optimization of the controls under the same terminal-cost objective. In temporal \gls{da}, however, intermediate noisy states do not provide direct access to future clean-state estimates such as $\E[\rvx_\tau\mid \rvz_t^{(s)}]$. As a result, optimizing each denoising-step control requires a full autoregressive rollout to the relevant observation time $\tau$, making \gls{ttoda} prohibitively expensive. This ablation isolates the benefit of amortizing controls through the offline-trained controller $\vu_\psi$. (ii) \textit{\gls{bon}} samples $n{=}16$ independent latent seeds and selects the trajectory with the lowest terminal cost, providing a simple inference-time selection baseline inspired by contemporary alignment and scaling approaches \citep{singhal2025general,gao2023scaling}.

\paragraph{Qualitative and Quantitative Analysis.}
Tab.~\ref{tab:rmsd_clean} compares \gls{cada} against state-of-the-art baselines and ablations. \gls{cada} performs best on most metrics and keeps \acrshort{rmse} nearly constant from short to long rollouts. The \acrshort{hct} results in Tab.~\ref{tab:hct_clean} (App.~\ref{app:hct}) show the same trend: \gls{cada} preserves high temporal correlations over full trajectories, whereas other \gls{ardm}-based methods, such as Plain and Universal amortized, degrade at long horizons and sometimes fail even in short rollouts. This stability is critical for operational \gls{da}, where forecasts must remain reliable over fixed-lag windows.

The factor-16 columns of Tab.~\ref{tab:rmsd_clean} and Tab.~\ref{tab:rmse-mr4} extend the standard linear operators to \acrshort{ds}-16 and \acrshort{ms}-16. Tab.~\ref{tab:nonlinear_preview} separates two further effects: robustness to a nonlinear observation map and sensitivity to the preview length $\Lambda$. These axes answer distinct questions. Factor 16 tests whether the controller can recover trajectories under an extreme spatial bottleneck; the nonlinear operator tests whether preview-aware control remains effective when observation compatibility is non-affine and sign-ambiguous; and $\Lambda{=}1$ removes anticipatory information while retaining the original sparse arrival process. All comparisons use the same rollout horizons and dataset-specific priors as the corresponding standard \acrshort{ms} experiments, so differences isolate the observation operator or preview horizon.

\begin{table}[t]
\centering
\setlength{\tabcolsep}{3.5pt}
\captionsetup{skip=4pt}
\caption{\textbf{\acrshort{cada} is both the most accurate and the fastest on \acrshort{era5}.} Left: \acrshort{rmse} under the \acrshort{ds}-$k$ and \textsc{\acrshort{ms}}-$k$ observation regimes defined in Sec.~\ref{par:obs-regimes}. Right: time to generate 8 trajectories (horizon 60) under \textsc{\acrshort{ms}}-8, in seconds and relative to \acrshort{cada}. Best in bold.}
\label{tab:rmse-mr4}
\label{tab:timing-mr8}

\footnotesize
\setlength{\tabcolsep}{2.5pt}
\begin{tabular}{@{}l cccccccc S[table-format=3.1] S[table-format=3.1]@{}}
  \toprule
  & \multicolumn{8}{c}{\textbf{\acrshort{rmse}$\downarrow$ on \acrshort{era5}}} & \multicolumn{2}{c}{\textbf{Time (\textsc{\acrshort{ms}-8})}} \\
  \cmidrule(lr){2-9}\cmidrule(lr){10-11}
  & \multicolumn{4}{c}{\acrshort{ds}} & \multicolumn{4}{c}{\acrshort{ms}} & & \\
  \cmidrule(lr){2-5}\cmidrule(lr){6-9}
   & \acrshort{ds}-2 & \acrshort{ds}-4 & \acrshort{ds}-8 & \acrshort{ds}-16 & \acrshort{ms}-2 & \acrshort{ms}-4 & \acrshort{ms}-8 & \acrshort{ms}-16 & {Time\,(s)} & {$\times$\,\acrshort{cada}} \\
  \midrule
  \textbf{\acrshort{cada} (ours)} & \textbf{0.68} & \textbf{0.75} & \textbf{0.99} & \textbf{0.07} & 1.23 & \textbf{1.32} & \textbf{1.43} & \textbf{0.13} & \bfseries 6.3 & \bfseries 1.0 \\
  Joint AAO & 0.77 & 1.75 & 2.70 & 3.16 & 2.10 & 2.31 & 3.49 & 4.36 & 326.8 & 51.9 \\
  Joint AR & 0.78 & 1.74 & 2.69 & 3.17 & \textbf{0.98} & 2.40 & 3.45 & 3.89 & 756.3 & 120.0 \\
  Plain amortized & 3.23 & 3.31 & 3.69 & 4.60 & 1.61 & 2.10 & 3.03 & 3.68 & 63.4 & 10.1 \\
  Universal amortized & 2.02 & 2.26 & 2.89 & 3.31 & 1.96 & 1.91 & 2.29 & 3.20 & 65.6 & 10.4 \\
  DiffDA & \NA & \NA & \NA & \NA & 1.41 & 1.65 & 2.25 & 3.64 & 29.6 & 4.7 \\
  \acrshort{ttoda} & 2.78 & 2.77 & 2.81 & 4.82 & 1.39 & 1.73 & 2.20 & 5.79 & 230.4 & 36.6 \\
  \acrshort{bon} & 3.27 & 3.96 & 3.97 & 5.96 & 3.91 & 3.91 & 4.34 & 6.23 & 25.8 & 4.1 \\
  \bottomrule

\end{tabular}
\end{table}

\begin{table*}[t]
\centering
\footnotesize
\setlength{\tabcolsep}{3.5pt}
\caption{\textbf{\acrshort{cada} stays accurate under nonlinear observations and with a one-step preview.} Left: \acrshort{rmse} under the nonlinear operator (NL-MS-$k$; Sec.~\ref{par:obs-regimes}), \acrshort{cada} vs.\ the best baseline; better in bold. Right: \acrshort{cada} with preview $\Lambda{=}1$ (default 16; 54 for \acrshort{ks}). Rollouts: 60 steps (640 for \acrshort{ks}).}
\label{tab:nonlinear_preview}
\begin{tabular}{ll cccc ccc}
\toprule
& & \multicolumn{4}{c}{\textbf{Nonlinear observations}} &
\multicolumn{3}{c}{$\boldsymbol{\Lambda=1}$} \\
\cmidrule(lr){3-6}\cmidrule(lr){7-9}
\textbf{Dataset} & \textbf{Method} &
\textbf{NL-MS-2} & \textbf{NL-MS-4} & \textbf{NL-MS-8} & \textbf{NL-MS-16} &
\textbf{MS-2} & \textbf{MS-4} & \textbf{MS-8} \\
\midrule
Kolmogorov & \textbf{\gls{cada} (ours)} & \textbf{0.029} & \textbf{0.041} & \textbf{0.058} & \textbf{0.297} & 0.041 & 0.041 & 0.095 \\
 & Best baseline & 0.046 & 0.144 & 0.316 & 0.320 & \NA & \NA & \NA \\
\acrshort{ks} & \textbf{\gls{cada} (ours)} & \textbf{0.008} & \textbf{0.019} & 0.847 & \textbf{1.094} & 0.009 & 0.021 & 0.024 \\
 & Best baseline & 0.017 & 0.039 & \textbf{0.154} & 1.182 & \NA & \NA & \NA \\
\acrshort{era5} & \textbf{\gls{cada} (ours)} & \textbf{1.003} & \textbf{1.885} & \textbf{2.309} & \textbf{2.329} & 1.312 & 1.327 & 1.507 \\
 & Best baseline & 1.760 & 2.154 & 2.571 & 3.496 & \NA & \NA & \NA \\
\bottomrule
\end{tabular}
\end{table*}

The non-amortized control ablation, \gls{ttoda}, performs worse than \gls{cada}, with the gap widening over long horizons (Kolmogorov \acrshort{ms}-4: 0.113/0.357 vs.\ 0.024/0.024 for short/long rollouts). This suggests that optimizing controls only at test time and arrival points is insufficient for stable assimilation; learning a reusable controller over observations and preview windows is essential. The \gls{bon} heuristic also underperforms (0.266/0.454 in the same setting): although best-of-$n$ selection can help static generation, \gls{da} requires consistent stepwise correction rather than ex-post selection. These results highlight amortized control as the key mechanism for sequential inverse problems.

Fig.~\ref{fig:compare} and Fig.~\ref{fig:ks} visualize long-horizon \gls{da} under \acrshort{ms}-4 observations for Kolmogorov flow (2D, horizon 180) and \gls{ks} (1D, horizon 640). Although trained with finite preview windows ($\Lambda{=}16$ and $54$, respectively), \gls{cada} remains stable on horizons more than $10\times$ longer than $\Lambda$.

\begin{figure}[t]
\centering
\includegraphics[width=\linewidth]{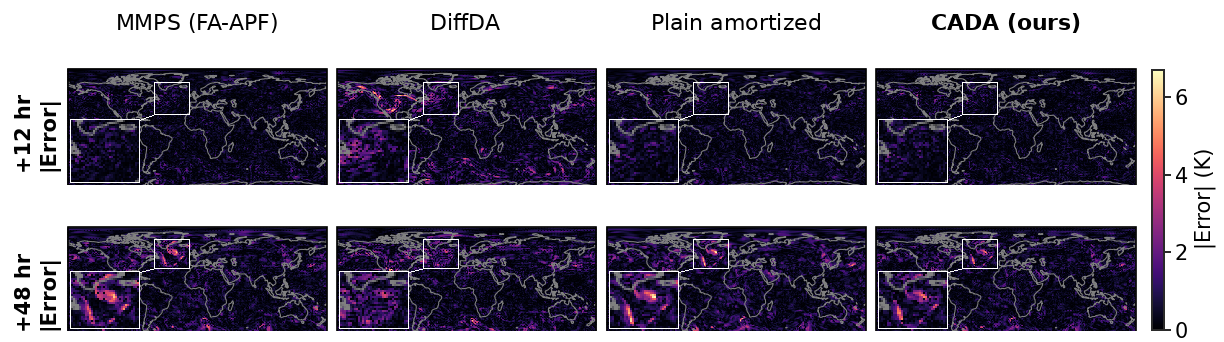}
\captionsetup{skip=3pt}
\captionof{figure}{\textbf{\acrshort{cada} keeps GenCast forecast errors small and localized.} Absolute 400 hPa temperature error (K) at 12 h (top) and 48 h (bottom) under \textsc{\acrshort{ms}}-4 observations; columns are methods. Insets magnify a North Atlantic storm whose error grows over time.}
\label{fig:gencast_temp400_main}
\vspace{3pt}
\begin{minipage}{\linewidth}
\centering
\captionsetup{skip=3pt}
\captionof{table}{\textbf{\acrshort{cada} attains the lowest GenCast temperature \acrshort{rmse} at every observation density.} Full-field \acrshort{rmse} averaged over the 4-step rollout (12--48 h leads), with observations at every step; rows are spatial strides (Sec.~\ref{par:obs-regimes}), columns are methods. Best in bold; the margin is largest at \textsc{\acrshort{ms}}-4.}
\label{tab:gencast_comparison_avg_full_rmse}
\footnotesize
\setlength{\tabcolsep}{6pt}
\renewcommand{\arraystretch}{1.1}
\begin{tabular}{lccccc}
\toprule
\textbf{Observation mask} & \textbf{Unguided} & \textbf{\begin{tabular}[b]{@{}c@{}}MMPS\\(FA-APF)\end{tabular}} & \textbf{DiffDA} & \textbf{\begin{tabular}[b]{@{}c@{}}Plain\\amortized\end{tabular}} & \textbf{\acrshort{cada} (ours)} \\
\midrule
Spatial stride 4 (\textsc{\acrshort{ms}}-4) & 0.77 & 0.73 & 0.91 & 0.77 & \textbf{0.65} \\
Spatial stride 8 (\textsc{\acrshort{ms}}-8) & 0.77 & 0.76 & 1.06 & 0.77 & \textbf{0.71} \\
Spatial stride 16 (\textsc{\acrshort{ms}}-16) & 0.78 & 0.77 & 0.90 & 0.78 & \textbf{0.74} \\
\bottomrule
\end{tabular}
\end{minipage}
\end{figure}

Tab.~\ref{tab:physics_diagnostics} reports these physics-aware errors under \acrshort{ms}-8 observations, over 640-step \gls{ks} and 60-step Kolmogorov rollouts. \acrshort{tv} measures the spatial roughness of the \gls{ks} field, so a low \acrshort{tv} error means the forecast keeps the ground truth's fine-scale structure instead of smoothing it out or adding spurious oscillations; \gls{cada}'s error (0.12) is $13\times$ lower than the next best (Joint AR, 1.56). The dissipation rate is a standard diagnostic of turbulent energy transfer to small scales; \gls{cada}'s error is $2\times$ lower than the next best (Plain amortized). Joint-score methods underestimate dissipation, while amortized \glspl{ardm} tend to overshoot it. Together, these results show that control augmentation not only reduces forecast drift but also improves domain-relevant physical fidelity.

In addition to improving accuracy and stability, \gls{cada} is substantially more efficient at inference. Tab.~\ref{tab:timing-mr8} reports wall-clock time for generating 8 \gls{era5} trajectories of 60 steps in the \textsc{\acrshort{ms}-8} regime. \gls{cada} is over $10\times$ faster than conditional \gls{ardm} baselines (Plain and Universal amortized) and $50$--$120\times$ faster than joint-score methods (Joint AAO/AR). This efficiency follows from using a single controlled rollout of the pretrained \gls{ardm}, whereas guidance-only and joint-score methods require repeated score or gradient evaluations during sampling.

We further validate \gls{cada} in three additional settings. First, we evaluate a compact \gls{era5} weather setup: we train our own \gls{ardm} from scratch on a $64{\times}64$ North America patch using only $(T,u,v)$ at 500 hPa from 2006--2016, and run the controller on top of this small prior. This shows that the controller transfers from synthetic \gls{pde}s to a realistic geophysical setting. Across most observation regimes, \gls{cada} achieves lower \acrshort{rmse} than conditional \gls{ardm}, joint-score baselines, and ablations (Tab.~\ref{tab:rmse-mr4}); Fig.~\ref{fig:era5vort} shows vorticity forecasts, and Fig.~\ref{fig:era5temp} (App.~\ref{app:addres}) the full comparison. Second, classical \gls{enkf}, \gls{3dvar}, and \gls{4dvar} are competitive only under dense downsampled observations (\gls{3dvar} beats \gls{cada} at \acrshort{ds}-8) but degrade sharply under masked, temporally strided ones, where \gls{cada}'s error is 5--15$\times$ lower (Tab.~\ref{tab:enkf}, Fig.~\ref{fig:enkf}; App.~\ref{app:addres}). Third, \gls{cada} also attains the lowest \acrshort{rmse} under randomized, non-grid-aligned spatial masks, without architectural changes (Tab.~\ref{tab:randmask}; App.~\ref{app:addres}).

\paragraph{Large-Scale GenCast Assimilation.}
\label{sec:gencast}

We additionally evaluate \gls{cada} in a larger-scale weather assimilation setting built on GenCast. Unlike in the compact \gls{era5} experiment, where we train a small prior from scratch, here the pretrained \gls{ardm} prior is GenCast \citep{price2025gencast} itself. The experiment uses $1^\circ$ atmospheric fields at 12-hour cadence. The controller is trained on February 2019 data and evaluated on January--March 2022. GenCast is rolled out over its full set of prognostic fields, while only temperature (all 13 pressure levels) is observed, through sparse masks.

\paragraph{Observation setting and baselines.}
We use spatial \textsc{\acrshort{ms}}-4/8 masked-observation regimes and include \textsc{\acrshort{ms}}-16 as an extreme-sparsity follow-up, with observations available at every 12-hour forecast step. \textsc{\acrshort{ms}}-4 follows the protocol of \citet{savary2025trainingfreedataassimilationgencast} (one grid point every $4^\circ$ in latitude and longitude); \textsc{\acrshort{ms}}-8 and \textsc{\acrshort{ms}}-16 use $8^\circ$ and $16^\circ$ spacing. Unlike the \gls{pde} \textsc{\acrshort{ms}} regimes, this isolates spatial sparsity: observations are not temporally strided. The controller is trained with 2-step preview windows and evaluated on 4-step rollouts, corresponding to forecast leads of 12, 24, 36, and 48 hours. We compare against unguided GenCast, DiffDA, Plain amortized guidance, and MMPS (FA-APF) \citep{savary2025trainingfreedataassimilationgencast}, which combines a Monte Carlo posterior sampler with a fully adapted auxiliary particle filter.

\paragraph{Results.}
Tab.~\ref{tab:gencast_comparison_avg_full_rmse} reports average full-field temperature \acrshort{rmse} over the 4-step rollout. \gls{cada} achieves the lowest error across all three spatial strides. For stride 4, it reduces average \acrshort{rmse} by about $11\%$ relative to MMPS (FA-APF) and $15\%$ relative to unguided GenCast; at stride 16, it improves on the next-best method by about $4.6\%$ (from unrounded values). MMPS (FA-APF) is consistently the strongest baseline, while Plain amortized remains close to unguided GenCast and DiffDA does not improve over the uncontrolled prior under this protocol. Although \gls{cada}'s margin narrows as the mask becomes sparser, it remains best at \textsc{\acrshort{ms}}-16. Fig.~\ref{fig:gencast_temp400_main} shows selected 400 hPa absolute-error maps; full forecast and error maps are in Fig.~\ref{fig:gencast_temp400} (App.~\ref{app:gencast-qualitative}). The errors for \gls{cada} remain more localized, indicating that the learned controller can scale to a high-dimensional weather prior while using sparse observations from a subset of fields.
\FloatBarrier
\section{Conclusion}
\label{sec:conclusion}

We presented \gls{cada}, an amortized control framework that recasts sparse-observation data assimilation as a single feed-forward rollout of a pretrained \gls{ardm}. A small, observation-aware controller is trained once to inject anticipatory corrections inside each denoising transition, replacing expensive test-time guidance, trajectory search, and adjoint-based optimization with a reusable policy.

Across two canonical chaotic \glspl{pde} and a compact \gls{era5} setup, \gls{cada} outperforms strong diffusion-based and classical \gls{da} baselines in most regimes, and most clearly under sparse, temporally strided observations, delivering more accurate forecasts, markedly better long-horizon stability, and closer agreement with domain-standard physical diagnostics. The same recipe applies without modification to a foundation-scale autoregressive weather model: a small controller trained on top of a frozen GenCast prior improves on both unguided GenCast and bespoke training-free \gls{da} methods designed for that prior (Sec.~\ref{sec:gencast}). Our ablations isolate amortization as the decisive ingredient: neither test-time optimization of the same objective nor best-of-$n$ trajectory selection recovers comparable robustness, indicating that what makes controlled diffusion practical here is not the control objective itself but the fact that its solution is learned offline and reused.

More broadly, our results point to a simple recipe for sequential inverse problems with generative priors: cast assimilation as control over a frozen \gls{ardm}, and amortize that control. The GenCast experiment suggests this recipe is compatible with off-the-shelf foundation models for the dynamics, opening a lightweight path to assimilating sparse, delayed, or noisy observations into priors that need no retraining, from operational weather and climate to robotics and scientific simulation.

\paragraph{Limitations and scope.}
We study two chaotic \glspl{pde}, compact \gls{era5}, and public GenCast, using one controller per observation regime. \gls{ardm} pretraining costs are comparable to existing diffusion-based \gls{pde} surrogates, while controllers add modest overhead and can use a frozen foundation prior. The GenCast study is limited to temperature observations and four-step forecasts, and therefore does not test longer operational cycling or heterogeneous observing systems. Moreover, each controller inherits the support and biases of its frozen prior: control can correct trajectories toward available observations, but cannot supply unresolved dynamics absent from the underlying model. Multivariate observations, cross-regime controllers, adaptive previews, and richer operators are natural extensions.

\clearpage
\begin{ack}
Stephan Mandt acknowledges funding from the National Science Foundation (NSF) through an NSF CAREER Award IIS-2047418, IIS-2611549, the NSF LEAP Center, the Chan Zuckerberg Initiative, SAP, and the Hasso Plattner Research Center at UCI.
\end{ack}

\bibliographystyle{unsrtnat}
\bibliography{citations}

\appendix

\newpage
\section{Observation Operators}
\label{app:obs-ops}

Our experiments use linear observation operators $A$ that map a full state $\rvx$ to an observed signal $\rvy$. These operators define the compatibility costs $\Phi(\rvx_\tau;\rvy_\tau)$ used at observation times in Eq.~\ref{eq:tilt_distribution}.

\paragraph{Masked observations.}
For spatially sparse observations, including temporally strided regimes when measurements arrive only at selected times, we use a binary mask $\mM \in \{0,1\}^{1\times D}$, where $D$ is the number of spatial grid points, broadcast across all channels of the state (so $d$ is $D$ times the number of channels):
\[
A_{\mathrm{mask}}(\rvx) \;=\; \mM \odot \rvx,
\qquad
\Phi^{\mathrm{mask}}(\rvx;\rvy) \;=\; 
\frac{\|\mM \odot \rvx - \rvy\|_2^2}{\|\mM\|_1}.
\]
The normalization by $\|\mM\|_1$ makes the cost comparable across different sparsity levels.

\paragraph{Downsampled observations.}
For coarse-resolution observations, we apply average pooling $P_f$ over non-overlapping $f{\times}f$ blocks, followed by nearest-neighbor upsampling $U_f$ back to the model grid:
\[
A_{\downarrow f}(\rvx) \;=\; U_f(P_f \rvx),
\qquad
\Phi^{\mathrm{ds}}(\rvx;\rvy) \;=\; 
\|U_f(P_f \rvx) - \rvy\|_2^2.
\]

These operators generate the observed signals $\rvy_\tau$ and the corresponding arrival-time costs used throughout the assimilation objective. Although our experiments focus on masking and downsampling, the framework can incorporate any differentiable observation operator and associated cost without modifying the controlled \gls{ardm}.

\section{Proof of the Tilted Distribution}
\label{app:proof-tilt}

We show that the optimization problem in Eq.~\ref{eq:arrival-cost} admits the tilted distribution in Eq.~\ref{eq:tilt_distribution} as its optimal solution.

\paragraph{Setup.}
Recall the objective
\[
\gC(\gP) \;=\; \sum_{\tau\in\gT} \E_{\rvx_\tau \sim \gP}\Big[\Phi(\rvx_\tau;\,\rvy_\tau)\Big] + \beta\,\KL(\gP \,\|\, \gQ),
\]
where $\gP$ is the guided \gls{cada} distribution over trajectories, $\gQ$ the unguided distribution, and $\Phi$ an arrival-time cost.

\paragraph{Variational form.}
Expanding the \gls{kl} divergence,
\[
\KL(\gP \,\|\, \gQ) \;=\; \E_{\gP}\!\left[ \log \frac{\gP}{\gQ} \right].
\]
Thus the objective reads
\[
\gC(\gP) \;=\; \E_{\gP}\!\Bigg[ \sum_{\tau\in\gT}\Phi(\rvx_\tau;\,\rvy_\tau) + \beta \log \frac{\gP}{\gQ}\Bigg].
\]

\paragraph{Lagrangian minimization.}
Consider minimizing $\gC(\gP)$ over distributions $\gP$ subject to normalization $\int \gP = 1$. The corresponding Lagrangian is
\[
\mathfrak{L}(\gP,\lambda) \;=\; \E_{\gP}\!\Bigg[ \sum_{\tau\in\gT}\Phi(\rvx_\tau;\,\rvy_\tau) + \beta \log \frac{\gP}{\gQ}\Bigg] + \lambda \Big(\int \gP - 1\Big).
\]

\paragraph{Stationary point.}
Taking the functional derivative w.r.t. $\gP$ gives
\[
\frac{\delta \mathfrak{L}}{\delta \gP} \;=\; \sum_{\tau\in\gT}\Phi(\rvx_\tau;\,\rvy_\tau) + \beta \Big(1 + \log \tfrac{\gP}{\gQ}\Big) + \lambda.
\]
Setting this derivative to zero yields
\[
\log \gP \;=\; \log \gQ - \tfrac{1}{\beta}\sum_{\tau\in\gT}\Phi(\rvx_\tau;\,\rvy_\tau) - \tfrac{\lambda+ \beta}{\beta}.
\]

\paragraph{Closed form.}
Exponentiating both sides gives
\[
\gP^* \;\propto\; \gQ \exp\!\Bigg(-\tfrac{1}{\beta}\sum_{\tau\in\gT}\Phi(\rvx_\tau;\,\rvy_\tau)\Bigg),
\]
which is exactly the tilted distribution in Eq.~\ref{eq:tilt_distribution}.

\section{Active Observation Selector}
\label{app:selector}

We maintain a preview buffer containing observations in $\mathcal{T}$ that fall within a preview window of length $\Lambda$ anchored at $t_0$. Each observation entry is represented by its observed signal and metadata, $(\rvy_j,\mM_j)$, where $j\in\mathcal{T}$ is the physical observation time, $\rvy_j$ is the observed signal lifted to the model resolution when needed, and $\mM_j$ is an auxiliary mask as in App.~\ref{app:obs-ops}. For non-masking operators, $\mM_j$ may be omitted or replaced by the corresponding operator metadata.

For transition $\rvx_{t-1}\to\rvx_{t}$, we define the set of future observations available within the anchored preview window as
\[
\mathcal{T}_{t}^{\Lambda}
\;\triangleq\;
\{\, j \in \mathcal{T} : t \le j \le t_0{+}\Lambda \,\}.
\]
The active preview is chosen as the nearest future observation in this set:
\[
\tau_t
=
\arg\min_{j \in \mathcal{T}_{t}^{\Lambda}} (j-t),
\qquad
\Delta_{t,\tau_t}= \tau_t - (t-1),
\]
and the preview context passed to the controller is
\[
\bar{\rvy}_{t}
=
(\rvy_{\tau_t},\, \mM_{\tau_t},\, \Delta_{t,\tau_t}).
\]
Thus, at each step, the selector provides the controller with the nearest previewed observation inside the current anchored window, together with its metadata and lead time. Details on how this selector is used during training and inference are given in App.~\ref{app:algo}. When $\mathcal T_t^\Lambda=\emptyset$, the selector returns a null preview context with empty metadata; the controller receives no active observation, and the observation cost is skipped until the next available arrival.

\section{Training and Sampling Algorithm}
\label{app:algo}

\begin{algorithm}[H]
\caption{Preview-aware controlled \acrshort{ddim} step ($\rvx_t \to \rvx_{t+1}$)}
\label{alg:controlled-step}
\DontPrintSemicolon
\SetAlgoNlRelativeSize{-1}
\SetInd{0.35em}{0.75em}

\KwIn{Current state $\rvx_t$; preview $\bar{\rvy}_{t+1}=(\rvy_{\tau_{t+1}},\mM_{\tau_{t+1}},\Delta_{t+1,\tau_{t+1}})$; pretrained \gls{ardm} kernels $q$; controller $\vu_\psi$; control scale $\gamma>0$}
\KwOut{Next state $\rvx_{t+1}$ and arrival-time cost $\ell_{t+1}$}

Sample parent latent $\rvz_{t+1}^{(S)} \sim q(\rvz_{t+1}^{(S)})$\;
$\ell_{t+1} \leftarrow 0$\;

\For{$s = S,S-1,\dots,1$}{
  $\rvu_{t+1}^{(s)} \leftarrow 
  \vu_\psi\!\left(\rvz_{t+1}^{(s)},\rvx_t,\bar{\rvy}_{t+1},s\right)$\;
  
  $\tilde{\rvz} \leftarrow \rvz_{t+1}^{(s)}+\gamma\,\rvu_{t+1}^{(s)}$\tcp*{latent shift}

  \If{$t+1\in\mathcal T$}{
    $\widehat{\rvx}_{t+1}^{(s)}
    \leftarrow
    \mathbb{E}\!\left[\rvz_{t+1}^{(0)}\mid \tilde{\rvz}\right]$\tcp*{Tweedie estimate}
    $\ell_{t+1} \leftarrow
    \ell_{t+1}
    +
    \Phi\!\left(\widehat{\rvx}_{t+1}^{(s)};\rvy_{t+1}\right)$\;
  }

  $\rvz_{t+1}^{(s-1)}
  \leftarrow
  \textsc{DenoiseStep}_{q}\!\left(\tilde{\rvz},\rvx_t,s\right)$\tcp*{\acrshort{ddim} step}
}

$\rvx_{t+1}\leftarrow \rvz_{t+1}^{(0)}$\;

\If{$t+1\in\mathcal T$}{
  $\ell_{t+1} \leftarrow \ell_{t+1}
  + \Phi(\rvx_{t+1};\rvy_{t+1})
  + \beta\,\KL$\tcp*{For the KL surrogate, see App.~\ref{app:kl_surrogate}}
}

\Return{$\rvx_{t+1},\ell_{t+1}$}\;
\end{algorithm}

\begin{algorithm}[H]
\caption{Training the controller network}
\label{alg:training}
\DontPrintSemicolon
\SetAlgoNlRelativeSize{-1}
\SetInd{0.35em}{0.75em}

\KwIn{Pretrained \gls{ardm} kernel $q$; observation stream $\{\rvy_j\}_{j\in\mathcal T}$; preview horizon $\Lambda$; regularization strength $\beta>0$; optimizer for $\psi$; controller $\vu_\psi$; control scale $\gamma>0$}
\KwOut{Trained controller parameters $\psi$}

\Repeat{convergence}{
  Sample rollout start $t_0$ and initial state $\rvx_{t_0}\sim p_0$\;
  $\mathcal T^\Lambda \leftarrow \mathcal T \cap [t_0+1,\,t_0+\Lambda]$\;
  $\widehat{\mathcal C} \leftarrow 0$\;

  \For{$t=t_0,t_0+1,\dots,t_0+\Lambda-1$}{
    $\mathcal T_{t+1}^\Lambda \leftarrow \{\,j\in\mathcal T: t+1\le j\le t_0+\Lambda\,\}$\tcp*{Anchored preview; App.~\ref{app:selector}}
    Build $\bar{\rvy}_{t+1}$ from $\mathcal T_{t+1}^\Lambda$\tcp*{App.~\ref{app:selector}}
    $(\rvx_{t+1},\ell_{t+1}) \leftarrow \textsc{ControlledStep}(\rvx_t,\bar{\rvy}_{t+1},q,\vu_\psi,\gamma)$\tcp*{Alg.~\ref{alg:controlled-step}}
    $\rvx_t \leftarrow \rvx_{t+1}$\;
    $\widehat{\mathcal C} \leftarrow \widehat{\mathcal C} + \ell_{t+1}\,\mathbf{1}\{t+1\in\mathcal T^\Lambda\}$\;
  }

  $\widehat{\mathcal C} \leftarrow \widehat{\mathcal C}/\max\{|\mathcal T^\Lambda|,1\}$\tcp*{Arrival normalization}
  $\mathcal L(\psi) \leftarrow \widehat{\mathcal C}$\;
  Update $\psi$ by descending $\nabla_\psi \mathcal L(\psi)$\;
}
\end{algorithm}

\begin{algorithm}[H]
\caption{Preview-aware forecasting with $\Lambda$-chunk anchoring}
\label{alg:sampling-chunked}
\DontPrintSemicolon
\SetAlgoNlRelativeSize{-1}
\SetInd{0.35em}{0.75em}

\KwIn{Pretrained \gls{ardm} kernel $q$; trained controller $\vu_\psi$; initial state $\rvx_{t_0}$; forecast horizon $H$; preview horizon $\Lambda$; observation stream $\{\rvy_j\}_{j\in\mathcal T}$; control scale $\gamma>0$}
\KwOut{Forecast trajectory $\rvx_{t_0+1:t_0+H}=(\rvx_{t_0+1},\dots,\rvx_{t_0+H})$}

$N \leftarrow \left\lceil H/\Lambda \right\rceil$\tcp*{Number of chunks}

\For{$c=0,1,\dots,N-1$}{
  $t_0^{(c)} \leftarrow t_0 + c\Lambda$\;
  $\Lambda_c \leftarrow \min\{\Lambda,\; H-c\Lambda\}$\tcp*{Last chunk may be shorter}

  \For{$t=t_0^{(c)},\dots,t_0^{(c)}+\Lambda_c-1$}{
    $\mathcal T_{t+1}^\Lambda \leftarrow \{\, j\in\mathcal T : t+1 \le j \le t_0^{(c)}+\Lambda_c \,\}$\tcp*{Anchored preview; App.~\ref{app:selector}}
    Build $\bar{\rvy}_{t+1}$ from $\mathcal T_{t+1}^\Lambda$\tcp*{App.~\ref{app:selector}}
    $(\rvx_{t+1},\,\ell_{t+1}) \leftarrow \textsc{ControlledStep}(\rvx_t,\bar{\rvy}_{t+1},q,\vu_\psi,\gamma)$\tcp*{Alg.~\ref{alg:controlled-step}}
  }

  \If{$c+1<N$}{
    $\rvx_{t_0^{(c+1)}} \leftarrow \rvx_{t_0^{(c)}+\Lambda_c}$\tcp*{Autoregressive handoff}
  }
}

\Return{$\rvx_{t_0+1:t_0+H}$}\;
\end{algorithm}

\section{\texorpdfstring{\gls{kl} regularization for guided \gls{ardm} training}{KL regularization for guided ARDM training}}
\label{app:kl_surrogate}

We give a tractable surrogate for the windowed \gls{kl} term
$\KL\!\big(\gP_\psi(\rvx_{t_0+1:t_0+\Lambda}\!\mid\!\rvx_{t_0}) \,\|\, \gQ(\rvx_{t_0+1:t_0+\Lambda}\!\mid\!\rvx_{t_0})\big)$. Throughout, $\gP_\psi$ is induced by the controlled kernel
(Eq.~\ref{eq:controlled-kernel}--\ref{eq:controlled-substep}) and $\gQ$ by the unguided \gls{ardm} (Eq.~\ref{eq:base-path}--\ref{eq:unguided_ardm}).

\paragraph{Path \gls{kl} and reduction to one-step \glspl{kl}.}
Conditioned on $\rvx_{t_0}$, both $\gP_\psi$ and $\gQ$ define Markov rollouts over $t=t_0,\dots,t_0+\Lambda-1$.
The chain rule for \gls{kl} therefore gives the exact decomposition

\begin{equation}
\label{eq:path_kl_exact}
\begin{aligned}
&\KL\!\Big(\gP_\psi(\rvx_{t_0+1:t_0+\Lambda}\mid \rvx_{t_0}) \,\big\|\, \gQ(\rvx_{t_0+1:t_0+\Lambda}\mid \rvx_{t_0})\Big) \\
&= \sum_{t=t_0+1}^{t_0+\Lambda} \E_{\rvx_{t-1}\sim \gP_\psi}\!\Big[ \KL\!\big(p(\rvx_{t}\mid \rvx_{t-1};\mU_{t}) \,\|\, q(\rvx_{t}\mid \rvx_{t-1})\big) \Big].
\end{aligned}
\end{equation}
The challenge is that $p(\rvx_{t}\mid \rvx_{t-1};\mU_{t})$ and $q(\rvx_{t}\mid \rvx_{t-1})$ are defined as marginals of
diffusion latent chains and the one-step \gls{kl} is generally intractable.

\paragraph{Augmenting with diffusion latents.}
Fix a physical transition $t{-}1\!\to\!t$ and introduce the joint latent chains
$\rvz_{t}^{(0:S)}$ with $\rvx_{t}\equiv \rvz_{t}^{(0)}$.
Define the corresponding augmented distributions
\begin{align}
\bar q(\rvz_{t}^{(0:S)}\mid \rvx_{t-1})
&\triangleq
q(\rvz_{t}^{(S)})\prod_{s=1}^{S} q(\rvz_{t}^{(s-1)}\mid \rvz_{t}^{(s)};\rvx_{t-1}),\\
\bar p_\psi(\rvz_{t}^{(0:S)}\mid \rvx_{t-1},\bar{\rvy}_{t})
&\triangleq
q(\rvz_{t}^{(S)})\prod_{s=1}^{S} p(\rvz_{t}^{(s-1)}\mid \rvz_{t}^{(s)};\rvu_{t}^{(s)},\rvx_{t-1}).
\end{align}
Since $\rvx_{t}$ is a marginal of $\rvz_{t}^{(0:S)}$, data processing yields the bound
\begin{equation}
\label{eq:dp_kl_bound}
\KL\!\big(p(\rvx_{t}\mid \rvx_{t-1};\mU_{t}) \,\|\, q(\rvx_{t}\mid \rvx_{t-1})\big)
\;\le\;
\KL\!\big(\bar p_\psi(\rvz_{t}^{(0:S)}\mid \rvx_{t-1},\bar{\rvy}_{t})\,\|\, \bar q(\rvz_{t}^{(0:S)}\mid \rvx_{t-1})\big).
\end{equation}
We use the latent-chain \gls{kl} on the right as a computable surrogate for the intractable marginal \gls{kl} on the left.

\paragraph{Sub-step \gls{kl} and closed form.}
Because $\bar p_\psi$ and $\bar q$ share the same prior $q(\rvz_{t}^{(S)})$, the joint \gls{kl} decomposes exactly along the reverse chain:
\begin{equation}
\label{eq:latent_kl_decomp}
\KL\!\big(\bar p_\psi \,\|\, \bar q\big)
=
\sum_{s=1}^{S}
\E_{\rvz_{t}^{(s)}\sim \bar p_\psi}\!\Big[
\KL\!\big(p(\rvz_{t}^{(s-1)}\mid \rvz_{t}^{(s)};\rvu_{t}^{(s)},\rvx_{t-1})\,\|\,q(\rvz_{t}^{(s-1)}\mid \rvz_{t}^{(s)};\rvx_{t-1})\big)
\Big].
\end{equation}
Under Eq.~\ref{eq:controlled-substep} and the corresponding unguided kernel, both sub-step conditionals are Gaussians with identical covariance
$\sigma_{s}^2\mI_d$, hence each \gls{kl} term admits the closed form
\begin{equation}
\label{eq:substep_kl_closed}
\KL(\cdot\|\cdot)
=
\frac{1}{2\sigma_{s}^2}
\big\|
\vmu(\rvz_{t}^{(s)}+\gamma \rvu_{t}^{(s)},\rvx_{t-1},s)
-
\vmu(\rvz_{t}^{(s)},\rvx_{t-1},s)
\big\|_2^2.
\end{equation}
Combining Eq.~\ref{eq:dp_kl_bound}--\ref{eq:substep_kl_closed} yields a tractable \gls{kl} regularizer that is fully aligned with our
guided-\gls{ardm} parameterization.

\paragraph{\gls{ddim} specialization.}
For \gls{ddim}, the per-substep \gls{kl} in Eq.~\ref{eq:substep_kl_closed} admits a standard
two-part form: a \emph{control-energy} term and a \emph{denoiser/score-discrepancy} term induced by the shift
$\rvz_{t}^{(s)}\mapsto \rvz_{t}^{(s)}+\gamma \rvu_{t}^{(s)}$, i.e.,
$\|\rvu_{t}^{(s)}\|_2^2$ and
$\|\epsilon(\rvz_{t}^{(s)}+\gamma \rvu_{t}^{(s)},\rvx_{t-1},s)-\epsilon(\rvz_{t}^{(s)},\rvx_{t-1},s)\|_2^2$,
where $\epsilon(\cdot)$ is the noise-prediction network underlying $\vmu$,
with step-dependent weights determined by the noise schedule.

\section{Data and Evaluation}
\label{app:expdet}

\paragraph{Datasets.}
The \gls{ks} equation is a fourth-order nonlinear \gls{pde} modeling flame-front instabilities and solidification dynamics:$\partial_t u + u\,\partial_\xi u + \partial_\xi^2 u + \nu\,\partial_\xi^4 u = 0,$ where $\nu>0$ is the viscosity. We solve it on a periodic domain with 256 spatial points and time step $\Delta t=0.2$. Training trajectories span $140\Delta t$, while validation and test trajectories extend to $640\Delta t$. Data generation and splits follow \citet{shysheya2024conditional, brandstetter2022lie}.

Kolmogorov flow is a two-dimensional incompressible Navier--Stokes system: $\partial_t \bm{v} + \bm{v}\cdot\nabla \bm{v} - \nu\nabla^2 \bm{v} + \tfrac{1}{\rho}\nabla p - f = 0,
\nabla\cdot \bm{v} = 0,$ with velocity field $\bm{v}=(v_\xi,v_\eta)$, viscosity $\nu$, density $\rho$, pressure $p$, and external forcing $f$. Trajectories contain 64 states for training and 180 states for validation and test, represented on a $64\times64$ grid with $\Delta t=0.2$. Data generation and splits follow \citet{shysheya2024conditional, rozet2023score}. We evaluate on the scalar vorticity field
$\Omega=\partial_\xi v_\eta-\partial_\eta v_\xi$, which captures rotational structures.

For a compact real-world \gls{da} benchmark, we use \gls{era5} reanalysis fields at 500\,hPa, extracting temperature and horizontal winds $(T,u,v)$ over a North America patch at 6-hour cadence (00/06/12/18 UTC). We use years 2006--2016, split by year, with the final year held out for validation/evaluation and the remaining years used for training. Each snapshot is interpolated to a $64\times64$ regular grid and partitioned into fixed-length 64-step trajectories. Per-channel min/max normalization statistics are computed on the training set only, and results are reported on all fields.

\paragraph{Evaluation Metrics.}
\acrshort{hct} is defined as the last index $m_{\max}$ for which the Pearson correlation $r(m)$ between forecast and ground truth remains above a fixed threshold $\phi$: $m_{\max} \;=\; \max \{m : r(m) \ge \phi \}, \quad t_{\max} = m_{\max}\,\Delta t.$ We use $\phi=0.9$, stricter than the $0.8$ of \citet{shysheya2024conditional}, so our \acrshort{hct} values are not directly comparable to theirs.

To assess physical fidelity beyond pointwise error, we also report domain-specific diagnostics. For the 1D \gls{ks} system, with state field $u(\xi)$ over spatial coordinate $\xi$, we compute the \acrshort{tv}, $\mathrm{TV}(u)
=
\int |\partial_\xi u(\xi)|\,d\xi,$ which measures spatial oscillation and the sharpness of evolving patterns. For 2D Kolmogorov flow, with velocity field $\bm{v}=(v_\xi,v_\eta)$ and scalar vorticity $\Omega=\partial_\xi v_\eta-\partial_\eta v_\xi$, we report the kinetic-energy dissipation rate $\varepsilon(\bm{v})
=
\nu \int \|\nabla \bm{v}(\xi,\eta)\|_F^2\,d\xi\,d\eta$. This diagnostic measures the rate at which kinetic energy is dissipated at small scales. For both diagnostics, Tab.~\ref{tab:physics_diagnostics} reports the error $e_D$ of Eq.~\ref{eq:physics-error}, which excludes the initial state.

\section{\texorpdfstring{Pretraining \gls{ardm}}{Pretraining ARDM}}
\label{app:ardm_details}

\paragraph{Implementation details.}
We adapt the 1D and 2D diffusion codebase from \texttt{lucidrains}\footnote{\url{https://github.com/lucidrains/denoising-diffusion-pytorch}} into an \gls{ardm} tailored for \gls{pde} forecasting. Each \gls{ardm} transition corresponds to one-step forecasting via a \gls{ddim} sampler with $S{=}3$ denoising steps, $v$-parameterization, and a sigmoid noise schedule.  

The backbone is a residual U-Net with multi-resolution attention and learned sinusoidal time embeddings:
\begin{verbatim}
dim = 64,
dim_mults = (1, 2, 4, 8),
learned_sinusoidal_dim = 128
\end{verbatim}
Attention layers are applied at intermediate and coarse resolutions, while residual blocks follow the standard Conv–Norm–SiLU design.

\paragraph{Training configuration.}
Models are trained on a single NVIDIA RTX A6000 GPU with mixed precision (\textsc{fp16}) and exponential moving average. The configuration is:  
\begin{verbatim}
train_batch_size = 32
train_lr = 3.2e-4
train_num_steps = 1000000
gradient_accumulate_every = 1
ema_decay = 0.995
ema_every = 10
\end{verbatim}

\section{Control Network Architecture}
\label{app:ctrl-net}

\paragraph{Overview.}
The \emph{controller} network $\vu_\psi$ produces  controls $\rvu_t^{(s)}$ used at each denoising sub-step. In the rollout, we write $\rvu_{t+1}^{(s)} = \vu_\psi(\cdot)$ for brevity; here we provide a detailed description of the architecture and conditioning.

\paragraph{Inputs.}
At each sub-step, the network receives five spatial tensors concatenated along channels:  
(i) the current latent $\rvz_{t+1}^{(s)}$,  
(ii) the previous state $\rvx_t$,  
(iii) the preview observation $\rvy_{\tau_t}$,  
(iv) the auxiliary mask $\mM_{\tau_t}$, and  
(v) the previous control $\rvu_{\mathrm{prev}}$.  
In addition, it conditions on scalar metadata: the preview lag $\Delta_{t,\tau_t}$, the local frame index $k$ within the preview window, and the current $\log$ of the \gls{snr} at diffusion sub-step $s$ from the \gls{ddim} schedule.

\paragraph{Backbone encoder.}
The concatenated inputs are passed through a shallow convolutional encoder with two $3{\times}3$ layers and group normalization. A downsample/upsample path provides limited multi-scale context: features are reduced to half resolution, then upsampled and fused back with the original resolution. A $1{\times}1$ fusion convolution followed by group normalization yields the encoded representation.

\paragraph{\gls{film} conditioning.}
Each scalar input is normalized to $[0,1]$ and embedded via a two-layer Multi-Layer Perceptron of hidden dimension \texttt{hid} (768 in all experiments). The three embeddings (lag, frame index, \gls{snr}) are concatenated and mapped to scale/shift coefficients $(a,b)$ through a linear layer. These coefficients modulate the encoded features in a \gls{film} style,
$
\text{feat} \;\mapsto\; \text{feat} \cdot (1+a) + b.
$

\paragraph{Residual head.}
A $3{\times}3$ convolutional head outputs the control increment $\delta_\psi$. This is added to a normalized copy of the previous control $\rvu_{\mathrm{prev}}$, producing
$
\rvu_t^{(s)} = \text{GroupNorm}(\rvu_{\mathrm{prev}}) + \delta_\psi.
$
At the first denoising step, $\rvu_{\mathrm{prev}}$ is set to zero.

\paragraph{Configuration.}
In our experiments we instantiate the control network as
$
\texttt{hid=768},
$
with group normalization (8 groups), hidden dimension $\texttt{hid}$ for the encoder, and \gls{film} embeddings of dimension $\texttt{hid}$. The architecture is sufficiently expressive to incorporate preview information into the denoising dynamics.

\paragraph{Training details.}
Gradients flow only into $\psi$ (the pretrained UNet is frozen).
We use gradient checkpointing at each UNet call and detach $\rvu_{\mathrm{prev}}$ within a frame to avoid deep denoising-step recurrences; memory scales with the number of checkpoints. Experiments work with $\beta=10^{-2}$ found via grid search while $\gamma=0.1$.

\section{HCT Metric}
\label{app:hct}

Tab.~\ref{tab:hct_clean} reports \gls{hct} for the two \gls{pde} benchmarks.

\begin{table}[ht]
\centering
\caption{\textbf{\acrshort{cada} stays correlated with the ground truth longest in nearly all regimes.} High-correlation time (\acrshort{hct}$\uparrow$: last step at which the Pearson correlation with ground truth is at least 0.9; App.~\ref{app:expdet}, capped at the rollout length) for Kolmogorov flow (short/long: 60/180 steps) and \acrshort{ks} (140/640) under \acrshort{ds}-$k$ and \acrshort{ms}-$k$ observations (Sec.~\ref{par:obs-regimes}). Rows: baselines and ablations (\acrshort{ttoda}: no amortization; \acrshort{bon}: best of 16 unguided samples). Best in bold, including ties.}
\label{tab:hct_clean}
\setlength{\tabcolsep}{4pt}
\footnotesize
\begin{tabular*}{\linewidth}{@{\extracolsep{\fill}}l *{12}{S[table-format=3.0]}}
\toprule
& \multicolumn{2}{c}{\textbf{\acrshort{ds}-2}} & \multicolumn{2}{c}{\textbf{\acrshort{ds}-4}} &
  \multicolumn{2}{c}{\textbf{\acrshort{ds}-8}} & \multicolumn{2}{c}{\textbf{\acrshort{ms}-2}} &
  \multicolumn{2}{c}{\textbf{\acrshort{ms}-4}} & \multicolumn{2}{c}{\textbf{\acrshort{ms}-8}} \\
\cmidrule(lr){2-3}\cmidrule(lr){4-5}\cmidrule(lr){6-7}\cmidrule(lr){8-9}\cmidrule(lr){10-11}\cmidrule(lr){12-13}
& {short} & {long} & {short} & {long} & {short} & {long} & {short} & {long} & {short} & {long} & {short} & {long} \\
\midrule
\multicolumn{13}{c}{\textbf{Kolmogorov flow}} \\
\midrule
\textbf{\acrshort{cada} (ours)}& \bfseries 60 & \bfseries 180 & \bfseries 60 & \bfseries 180 & \bfseries 50 & 50 & \bfseries 60 & \bfseries 180 & \bfseries 60 & \bfseries 180 & \bfseries 60 & \bfseries 70 \\
Joint AAO                     & \bfseries 60 & \bfseries 180 & \bfseries 60 & \bfseries 180 & \bfseries 50 & \bfseries 180 & \bfseries 60 & 173 & 25 & 25 & 10 & 18 \\
Joint AR                      & \bfseries 60 & \bfseries 180 & \bfseries 60 & \bfseries 180 & \bfseries 50 & \bfseries 180 & \bfseries 60 & 177 & \bfseries 60 & 62 & 13 & 13 \\
Plain amortized               & \bfseries 60 & 64 & 41 & 45 & 13 & 17 & 38 & 42 & 32 & 33 & 31 & 32 \\
Universal amortized           & 37 & 13 & 17 & 17 & 8 & 8 & 50 & 50 & 32 & 32 & 28 & 28 \\
DiffDA                        & \NA & \NA & \NA & \NA & \NA & \NA & 38 & 39 & 31 & 33 & 27 & 28 \\
\acrshort{ttoda}              & \bfseries 60 & 73 & \bfseries 60 & \bfseries 180 & 37 & 40 & \bfseries 60 & 60 & 48 & 40 & 27 & 25 \\
\acrshort{bon}                & 40 & 40 & 32 & 26 & 21 & 26 & 40 & 26 & 32 & 26 & 32 & 27 \\
\midrule
\multicolumn{13}{c}{\textbf{\acrlong{ks}}} \\
\midrule
\textbf{\acrshort{cada} (ours)}& \bfseries 140 & \bfseries 640 & \bfseries 140 & \bfseries 640 & \bfseries 140 & \bfseries 640 & \bfseries 140 & \bfseries 640 & \bfseries 140 & \bfseries 640 & \bfseries 140 & \bfseries 640 \\
Joint AAO                     & \bfseries 140 & \bfseries 640 & 139 & \bfseries 640 & 139 & \bfseries 640 & \bfseries 140 & \bfseries 640 & \bfseries 140 & \bfseries 640 & 129 & 633 \\
Joint AR                      & 139 & 639 & 139 & 639 & 139 & 639 & 139 & 639 & 139 & 639 & 139 & 639 \\
Plain amortized               & \bfseries 140 & 211 & \bfseries 140 & 147 & 76 & 56 & \bfseries 140 & 253 & \bfseries 140 & 264 & \bfseries 140 & 242 \\
Universal amortized           & \bfseries 140 & 262 & \bfseries 140 & 168 & 62 & 62 & \bfseries 140 & 286 & \bfseries 140 & 217 & \bfseries 140 & 274 \\
DiffDA                        & \NA & \NA & \NA & \NA & \NA & \NA & \bfseries 140 & \bfseries 640 & \bfseries 140 & 638 & \bfseries 140 & 627 \\
\acrshort{ttoda}              & \bfseries 140 & 218 & \bfseries 140 & \bfseries 640 & 139 & 633 & \bfseries 140 & 638 & \bfseries 140 & 637 & \bfseries 140 & 634 \\
\acrshort{bon}                & \bfseries 140 & 260 & \bfseries 140 & 250 & \bfseries 140 & 240 & \bfseries 140 & 243 & \bfseries 140 & 250 & \bfseries 140 & 246 \\
\bottomrule
\end{tabular*}
\end{table}

\section{Additional Results}
\label{app:addres}

\begin{figure}[ht]
  \centering
  \includegraphics[width=0.6\linewidth]{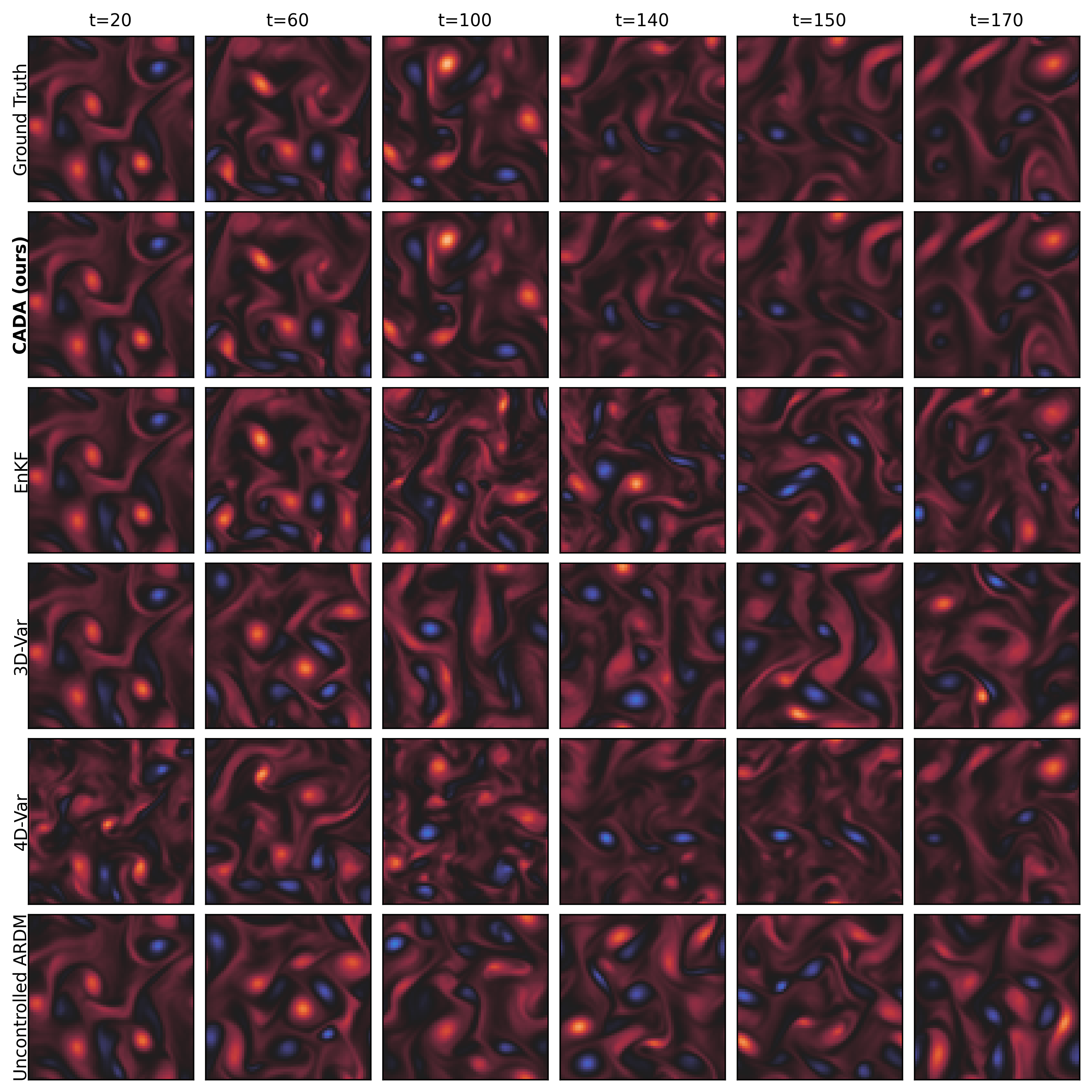}
  \caption{
  \textbf{\acrshort{cada} stays aligned with the ground truth over 170 steps, while classical \acrshort{da} methods and the uncontrolled \acrshort{ardm} drift.}
  Kolmogorov vorticity under \acrshort{ms}-4 observations; columns are rollout times from $t{=}20$ to $t{=}170$, and rows show ground truth, \acrshort{cada}, \acrshort{enkf}, \acrshort{3dvar}, \acrshort{4dvar}, and the uncontrolled \acrshort{ardm}. \acrshort{rmse} at 60 steps in Tab.~\ref{tab:enkf}.
  }
    \label{fig:enkf}
\end{figure}

To contextualize our method relative to classical techniques, we additionally evaluate standard \gls{enkf}, \gls{3dvar}, and \gls{4dvar} implementations (Tab.~\ref{tab:enkf}). With downsampled observations at every step, classical methods remain accurate (\acrshort{rmse} at most 0.19, except \gls{enkf} at \acrshort{ds}-8), and \gls{3dvar} even beats \gls{cada} at \acrshort{ds}-8 (0.09 vs.\ 0.138). Under masked, temporally strided observations, all three degrade sharply (0.26--0.42), whereas \gls{cada} stays at 0.017--0.060, 5--15$\times$ lower than the best classical method. This behavior is expected: linear–Gaussian assumptions and reliance on second-order statistics limit \gls{enkf}'s ability to recover fine-scale nonlinear structures in chaotic \gls{pde} systems, whereas \gls{cada} combines an expressive diffusion prior with amortized control for non-Gaussian, intermittently observed dynamics. Fig.~\ref{fig:enkf} shows a qualitative comparison.

\begin{figure}[t]
  \centering
  \begin{minipage}[t]{0.35\linewidth}
    \centering
    \includegraphics[width=\linewidth]{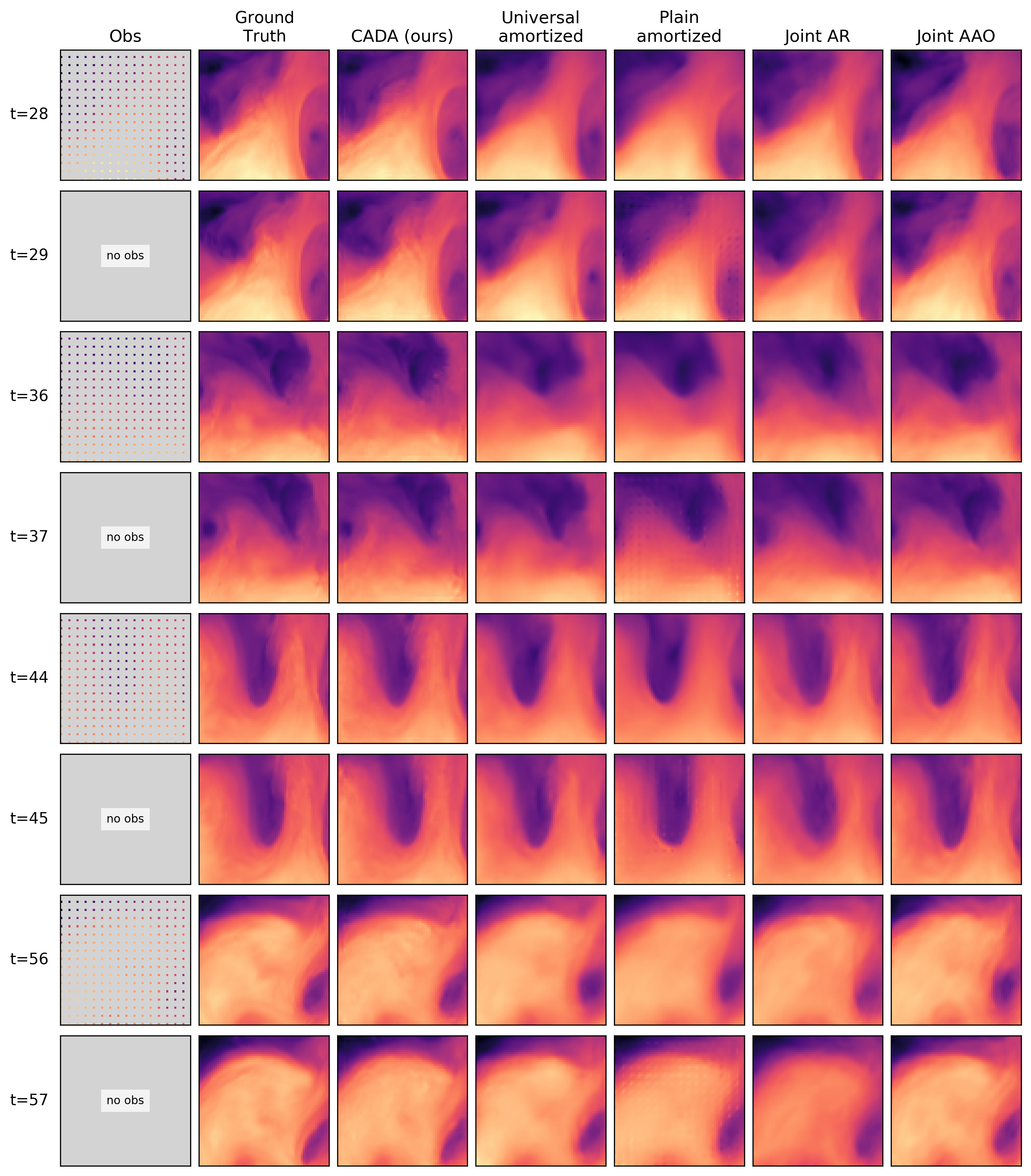}
  \end{minipage}\hfill
  \begin{minipage}[t]{0.35\linewidth}
    \centering
    \includegraphics[width=\linewidth]{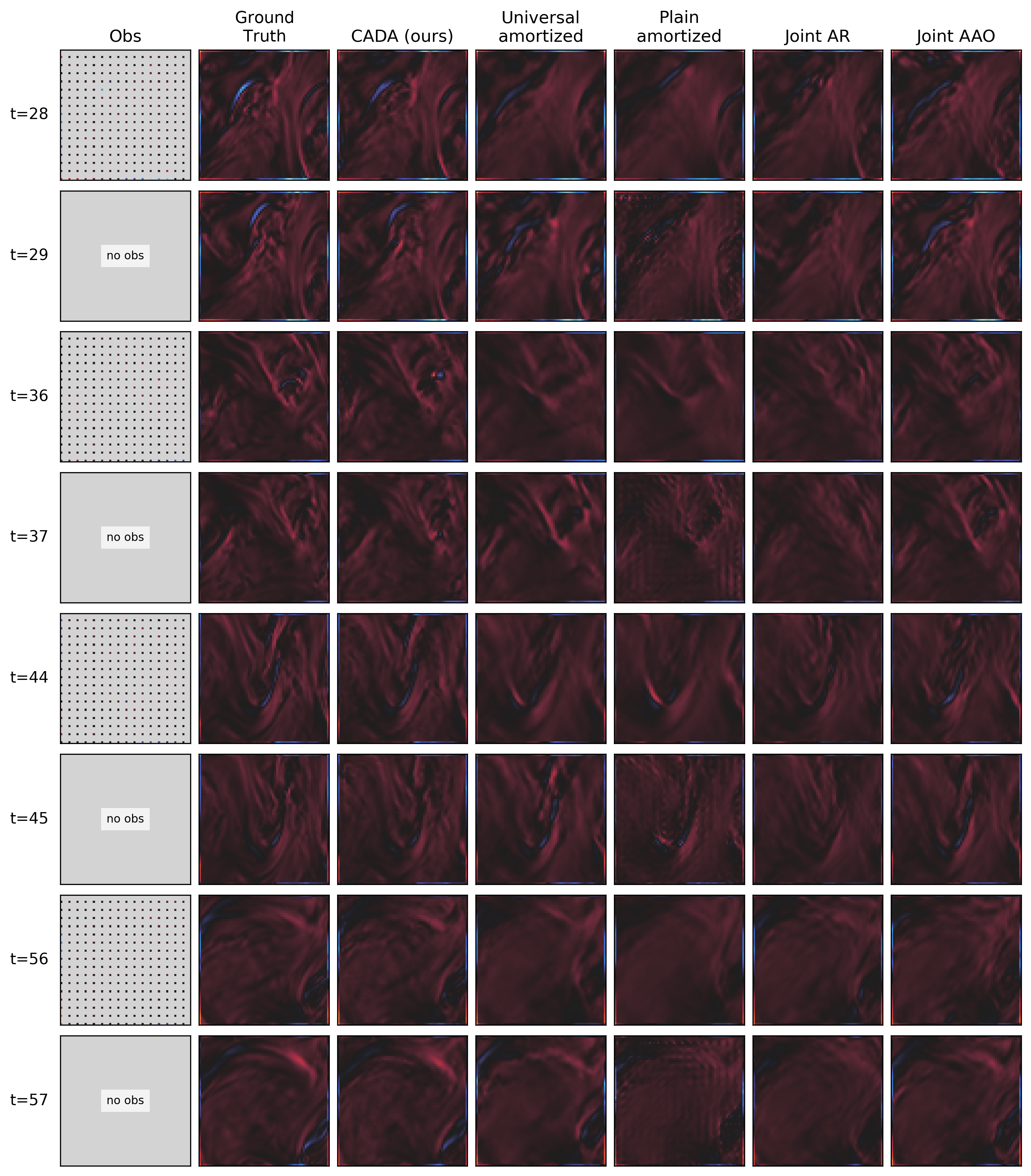}
  \end{minipage}
  \caption{
    \textbf{\acrshort{cada} keeps sharper temperature fronts and more coherent vorticity filaments on compact \acrshort{era5}, at observed and unobserved steps alike.} 500 hPa temperature (left) and vorticity (right) over North America under \acrshort{ms}-4 observations; rows are forecast times, and columns show observations (gray: none; dots: sparse measurements), ground truth, and reconstructions by \acrshort{cada}, Universal amortized, Plain amortized, Joint AR, and Joint AAO. Full comparison for Fig.~\ref{fig:era5vort}; \acrshort{rmse} in Tab.~\ref{tab:rmse-mr4}.}
  \label{fig:era5temp}
\end{figure}

\begin{table*}[t]
\centering

\begin{minipage}[t]{0.48\textwidth}
\centering
\captionof{table}{\textbf{\acrshort{cada} is the most accurate under irregular, randomly masked observations.} \acrshort{rmse}$\downarrow$ on Kolmogorov flow when each grid point is observed independently with probability $p=0.125$, at irregular times 2--6 steps apart; rows are methods. Best in bold.}
\label{tab:randmask}
\footnotesize
\begin{tabular}{lc}
\toprule
\textbf{Method} & \textbf{\acrshort{rmse}} $\downarrow$ \\
\midrule
Plain amortized      & 0.28 \\
Universal amortized  & 0.26 \\
Joint AAO            & 0.18 \\
Joint AR             & 0.07 \\
DiffDA & 0.23 \\
\acrshort{ttoda}               & 0.13 \\
\acrshort{bon}                  & 0.32 \\
\textbf{\acrshort{cada} (ours)} & \textbf{0.02} \\
\bottomrule
\end{tabular}
\end{minipage}
\hfill
\begin{minipage}[t]{0.48\textwidth}
\centering
\captionof{table}{\textbf{Classical \acrshort{da} methods do best with downsampled observations at every step and degrade sharply under masked, strided ones.} \acrshort{rmse}$\downarrow$ on Kolmogorov flow (60-step rollouts); rows are observation regimes (Sec.~\ref{par:obs-regimes}), columns are \acrshort{enkf}, \acrshort{3dvar}, \acrshort{4dvar}, and \acrshort{cada} (short-rollout values from Tab.~\ref{tab:rmsd_clean}). Best in bold.}
\label{tab:enkf}
\footnotesize
\begin{tabular}{lcccc}
\toprule
\textbf{Regime} & \textbf{\acrshort{enkf}} & \textbf{\acrshort{3dvar}} & \textbf{\acrshort{4dvar}} & \textbf{\acrshort{cada}} \\
\midrule
\acrshort{ds}-2 & 0.08 & 0.08 & 0.06 & \textbf{0.016} \\
\acrshort{ds}-4 & 0.09 & 0.11 & 0.18 & \textbf{0.020} \\
\acrshort{ds}-8 & 0.37 & \textbf{0.09} & 0.19 & 0.138 \\
\acrshort{ms}-2 & 0.31 & 0.38 & 0.26 & \textbf{0.017} \\
\acrshort{ms}-4 & 0.32 & 0.41 & 0.28 & \textbf{0.024} \\
\acrshort{ms}-8 & 0.35 & 0.42 & 0.33 & \textbf{0.060} \\
\bottomrule
\end{tabular}
\end{minipage}

\end{table*}

Lastly, assessing robustness beyond structured downsampling and regular masking, we conduct an additional experiment using \emph{irregular} spatial observations. Each grid point is independently revealed with probability $p=0.125$, producing non-aligned, non-regular measurement patterns. Observations follow an irregular cadence, with a minimum separation of 2 time steps and a maximum of 6 time steps between consecutive observations. As shown in Tab.~\ref{tab:randmask}, \gls{cada} attains the lowest \acrshort{rmse} (0.02 vs.\ 0.07 for the next best, Joint AR), while conditional \gls{ardm} baselines experience substantial degradation. These results reinforce that preview-based amortized control remains stable even when observations deviate significantly from regular masks, and that no architectural changes are required to accommodate such irregular regimes.

\clearpage
\section{Qualitative GenCast Results}
\label{app:gencast-qualitative}
\begin{figure}[ht]
    \centering
    \includegraphics[width=\linewidth]{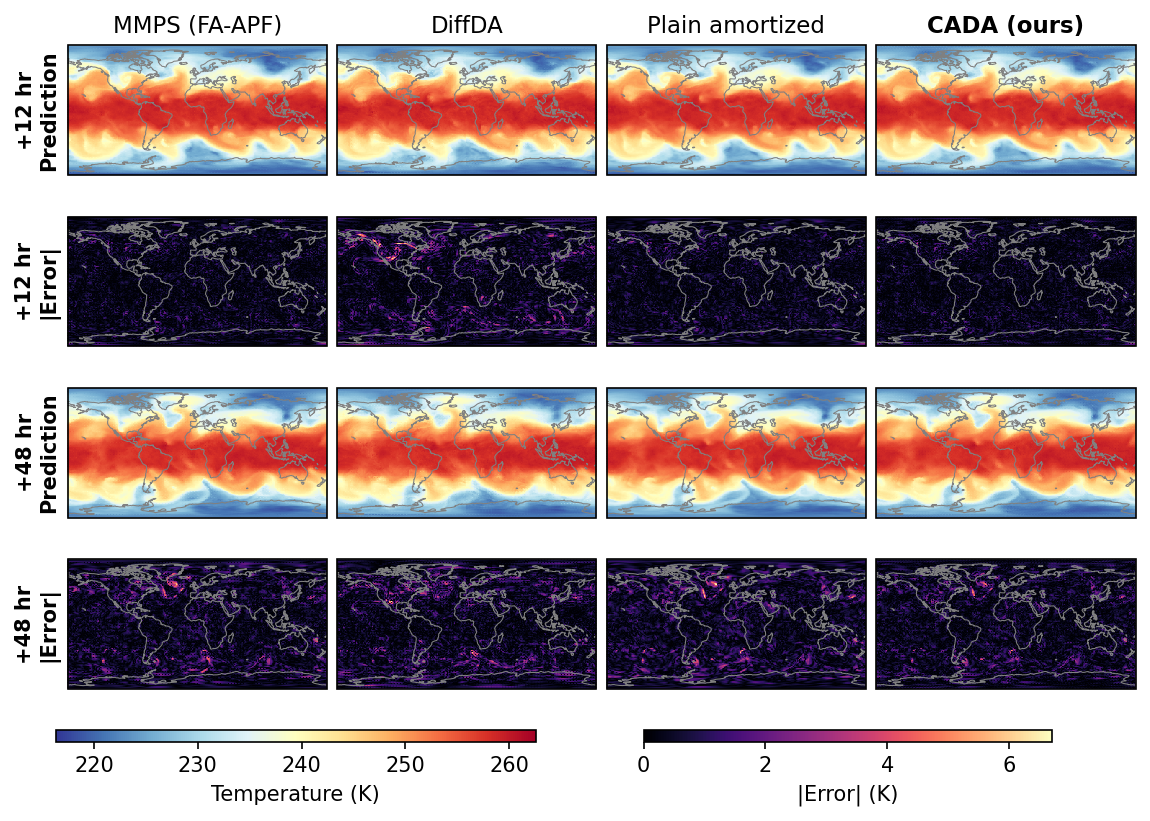}
    \caption{\textbf{\acrshort{cada} keeps GenCast temperature errors lower and more localized over 48 hours.} Global 400 hPa temperature under \textsc{\acrshort{ms}}-4 observations; columns are methods (MMPS (FA-APF), DiffDA, Plain amortized, \acrshort{cada}), and rows show the forecast (K) and its absolute error (K) at 12 h and 48 h leads. Full maps for Fig.~\ref{fig:gencast_temp400_main}.}
\label{fig:gencast_temp400}
\end{figure}

\end{document}